\documentclass[conference, 11pt]{IEEEtran}
\ifCLASSINFOpdf
\else
\fi
\usepackage{graphicx}
\usepackage{amsmath}
\usepackage{url}
\usepackage{color}
\usepackage{subfig}

\begin{document}


%\title{Layer-wise spiking convolutional neural networks trained by STDP-based representation learning and backpropagation algorithms}

\title{Deep Learning in Spiking Neural Networks}



\author{\IEEEauthorblockN{Amirhossein Tavanaei\IEEEauthorrefmark{1},
Masoud Ghodrati\IEEEauthorrefmark{2},
Saeed Reza Kheradpisheh\IEEEauthorrefmark{3}, \\ 
Timoth\'{e}e Masquelier\IEEEauthorrefmark{4} and
Anthony Maida\IEEEauthorrefmark{1}}
\\

\IEEEauthorblockA{\IEEEauthorrefmark{1}Center for Advanced Computer Studies, University of Louisiana at Lafayette \\ Lafayette, Louisiana, LA 70504, USA}
\IEEEauthorblockA{\IEEEauthorrefmark{2}Department of Physiology, Monash University, Clayton, VIC, Australia}
\IEEEauthorblockA{\IEEEauthorrefmark{3}Department of Computer Science, Faculty of Mathematical Sciences and Computer,\\ Kharazmi University, Tehran, Iran}
\IEEEauthorblockA{\IEEEauthorrefmark{4}CERCO UMR 5549,
CNRS-Universit\'{e} de Toulouse 3,
F-31300, France}
\\

\IEEEauthorblockA{\texttt{tavanaei@louisiana.edu,  masoud.ghodrati@monash.edu,
kheradpisheh@ut.ac.ir,}\\
\texttt{timothee.masquelier@cnrs.fr,
maida@louisiana.edu}}
}


\maketitle

\begin{abstract}
\color{red}
\footnote{\textbf{The final/complete version of this paper has been published in the Neural Networks journal. Please cite as:} \textit{Tavanaei, A., Ghodrati, M., Kheradpisheh, S. R., Masquelier, T., and  Maida, A. (2018). Deep learning in spiking neural networks. Neural Networks.} }
\color{black}
In recent years, deep learning has revolutionized
%been a revolution in 
the field of machine learning, for computer vision in particular. In this approach, a deep (multilayer) artificial neural network (ANN) is trained in a supervised manner using backpropagation. 
%Huge 
Vast amounts of labeled training examples are required, but the resulting classification accuracy is truly impressive, sometimes outperforming humans. Neurons in an ANN are characterized by a single, static, continuous-valued activation. Yet biological neurons use discrete spikes to compute and transmit information, and the spike times, in addition to the spike rates, matter. Spiking neural networks (SNNs) are thus more biologically realistic than ANNs, and arguably the only viable option if one wants to understand how the brain computes. SNNs are also more hardware friendly and energy-efficient than ANNs, and are thus appealing for technology, especially for portable devices. However, training deep SNNs remains a challenge. Spiking neurons' transfer function is usually non-differentiable, which prevents using backpropagation. Here we review recent supervised and unsupervised methods to train deep SNNs, and compare them in terms of accuracy, but also computational cost and hardware friendliness. The emerging picture is that SNNs still lag behind ANNs in terms of accuracy, but the gap is decreasing, and can even vanish on some tasks, while SNNs typically require many fewer operations.
\linebreak

\textbf{Keywords:} \textit{Deep learning, Spiking neural network, Biological plausibility, Machine learning, Power-efficient architecture.}
\end{abstract}


% \fi
\IEEEpeerreviewmaketitle

\section{Introduction}
Artificial neural networks (ANNs) are predominantly built using
idealized computing units with continuous activation values and
a set of weighted inputs.
These units are commonly called `neurons' because of their biological inspiration.
These (non-spiking) neurons use differentiable, non-linear activation functions.
The non-linear activation functions make it representationally meaningful to stack more than one
layer 
and the existence of their derivatives makes it possible to use gradient-based
optimization methods for training.
With recent advances in availability of large labeled data sets, computing power in the form of
general purpose GPU computing, and advanced regularization methods, these networks have become
very deep (dozens of layers) with great ability to generalize to unseen data and there have been huge advances in
the performance of such networks.

A distinct historical  landmark is the 2012 success of AlexNet~\cite{krizhevsky2012} in the ILSVRC image classification challenge~\cite{ILSVRC15}.
AlexNet became known as a deep neural network (DNN) because it consisted of about eight sequential layers of end-to-end learning,
totaling 60 million trainable parameters.
For recent reviews of DNNs, see \cite{lecun2015deep,schmidhuber2015}. DNNs have been remarkably successful in many applications including image recognition~\cite{krizhevsky2012,szegedy2016rethinking,he2016deep}, object detection~\cite{long2015fully,girshick2014rich}, speech recognition~\cite{hinton2012deep}, 
biomedicine and bioinformatics~\cite{mamoshina2016applications,min2017deep}, temporal data processing~\cite{venna2017novel}, 
and many other applications~\cite{schmidhuber2015,hassabis2017neuroscience,vanrullen2017perception}.  
These recent advances in artificial intelligence (AI) have opened up new avenues for developing different engineering applications and understanding of how biological brains work~\cite{hassabis2017neuroscience,vanrullen2017perception}.

Although DNNs are historically brain-inspired, 
there are fundamental differences in their structure, neural computations, and learning rule compared to the brain.
One of the most important differences is the way that information propagates between their units. 
It is this observation that leads to the realm of spiking neural networks (SNNs).
In the brain, the communication between neurons is done by broadcasting trains of action potentials, also known as spike trains to downstream neurons. 
These individual spikes are sparse in time, so each spike has high information content, and to a first approximation
has uniform amplitude (100 mV with spike width about 1 msec).
Thus, information in 
%spiking neural networks
SNNs is conveyed by spike timing, including latencies, and spike rates, possibly over populations~\cite{gerstner2014neuronal}.
SNNs almost universally use idealized spike generation mechanisms in contrast to the actual biophysical
mechanisms~\cite{hodgkin1952quantitative}.
 
ANNs, that are
non-spiking DNNs, communicate using continuous valued activations. Although the energy efficiency of DNNs can likely be improved,
SNNs offer a special opportunity in this regard because,
as explained below, spike events are sparse in time. Spiking networks also have the advantage of being intrinsically sensitive to the temporal characteristics of information  transmission that occurs in the biological neural systems. It has been shown that the precise timing of every spike is highly reliable for several areas of the brain 
and suggesting an important role in neural coding \cite{Sejnowski1995a,Bair1996a,Herikstad2011a}. 
This precise temporal pattern in spiking activity is considered as a crucial coding strategy in sensory information processing areas~\cite{gollisch2008rapid,sinha2017cellular,victor2005spike,butts2007temporal,reinagel2000temporal} and neural motor control areas in the brain~\cite{srivastava2017motor,tang2014millisecond}. 
%The biologically plausible artificial neural networks that take the time and spike-based information transformation into account are called spiking neural networks (SNNs).
SNNs have become the focus of a number of recent applications in many areas of pattern recognition such as visual processing~\cite{wysoski2010,gupta2007,meftah2010,escobar2009}, speech recognition~\cite{liaw1998,kroger2009,panchev2004,namarvar2001,wade2010,nager2002,loiselle2005},  and medical diagnosis~\cite{ghosh2007,kasabov2014b}. In recent years, a new generation of neural networks that incorporates the multilayer structure of DNNs (and the brain) and the type of information communication in SNNs has emerged. These deep SNNs are great candidates to investigate neural computation and different coding strategies in the brain.

In regard to the scientific motivation, it is well accepted that the ability of the brain to recognize complex visual patterns or identifying auditory targets in a noisy environment is a result of several processing stages and multiple learning mechanisms embedded in deep spiking networks~\cite{felleman1991distributed,serre2014hierarchical,freiwald2010functional}. In comparison to traditional deep networks, training deep spiking networks is in its early phases. 
It is an important scientific question to understand how such networks can be trained to perform different tasks as this can help us to generate and investigate  novel  hypotheses, 
such as rate versus temporal coding, and develop experimental ideas prior to performing physiological experiments. In regard to the engineering motivation, SNNs have some advantages over traditional neural networks in regard to implementation in special purpose hardware. At the present, effective training of traditional deep networks requires the use of energy intensive high-end graphic cards. 
Spiking networks have the interesting property that the output spike trains can be made sparse in time. 
An advantage of this in biological networks is that the spike events consume energy and that using few spikes which have high information content reduces energy 
consumption \cite{Stone2018a}. 
This same advantage is maintained in hardware~\cite{merolla2014million,seo201145nm,carrillo2012advancing,carrillo2013scalable}. 
Thus, it is possible to create low energy spiking hardware based on the property that spikes are sparse in time.

An important part of the learning in deep neural models, both spiking and non-spiking, 
occurs in the feature discovery hierarchy, where increasingly complex, discriminative, abstract, and invariant features are acquired~\cite{bengio2009learning}. 
Given the scientific and engineering motivations mentioned above, deep SNNs provide appropriate architectures for developing an efficient, brain-like representation.
Also, pattern recognition in the primate's brain is done through multi-layer neural circuits that communicate by spiking events. 
This naturally leads to interest in using artificial SNNs in applications that brains are good at, such as pattern recognition~\cite{maass2015spike}.
%There has been much research on this topic with notable successes.
 Bio-inspired SNNs, in principle, have higher representation power and capacity than traditional rate-coded networks~\cite{maass1997}. Furthermore, SNNs allow a type of bio-inspired learning (weight modification) that depends on the relative timing of spikes between pairs of directly connected neurons in which the information required for weight modification is locally available. 
This local learning resembles the remarkable learning that occurs in many areas of the brain~\cite{rozenberg2011handbook,seung2003learning,liu2005repeated,song2013asynchronous,chavez1990decrease}.
 
The spike trains are represented formally by sums of Dirac delta functions and do not have derivatives.
%The biologically realistic spiking neurons communicate using spike trains which do not have obvious derivatives. 
This makes it difficult to use derivative-based optimization for training SNNs, 
although very recent work has explored the use of various types of substitute or approximate derivatives~\cite{huh2017gradient,lee2016training}. 
This raises a question: How are neural networks in the brain trained if derivative-based optimization is not available?
Although spiking networks 
have theoretically been shown to have Turing-equivalent computing power~\cite{Maass1996a},
%considered to have higher computational power than rate-coded networks~\cite{maass2015spike}, 
it remains a challenge to train SNNs, especially deep SNNs using multi-layer learning. 
In many existing spiking networks, learning is restricted to a single layer, for
example~\cite{masquelier2007,tavanaei2016c,beyeler2013}. 
Equipping spiking networks with multi-layer learning is an open area that has potential to greatly improve their performance on different tasks.
The main core of the previous research is based on the fact that coding with the timing of spikes carries useful information and exhibits great computational power in biological systems~\cite{butts2007temporal,reinagel2000temporal,srivastava2017motor}. 

Here, we review recent studies in developing deep learning models in SNNs with the focus on:
(1) describing the SNNs' architectures and their learning approaches;
(2) reviewing deep SNNs composed of feedforward, fully connected spiking neural layers;
(3) spiking convolutional neural networks;
(4) reviewing spiking restricted Boltzmann machines and spiking deep belief networks; 
(5) reviewing recurrent SNNs; and
(6) providing a comprehensive summary comparing the performance of recent deep spiking networks. We hope that this review will help researchers 
in the area of artificial neural networks to develop and extend efficient and high-performance deep SNNs and will also foster a cross-fertilization in future experimental and theoretical work in neuroscience.

% Section 2
\section{Spiking Neural Network: A Biologically Inspired Approach to Information Processing}
The introduction of SNNs in the last few decades, as a powerful third generation neural network~\cite{maass1997}, has encouraged many
% a number of 
studies with the focus on biologically motivated approaches for pattern recognition~\cite{ghosh2009b,kasabov2013}. 
SNNs were originally inspired by the brain and the communication scheme that neurons use for information transformation via discrete action potentials (spikes) in time through adaptive synapses. 
In a biological neuron, a spike is generated when the running sum of changes in the membrane potential, which can result from presynaptic stimulation, crosses a threshold.
The rate of spike generation and the temporal pattern of spike trains carry information about external stimuli~\cite{gerstner2002spiking,rieke1999spikes} 
and ongoing calculations. 
SNNs use a very similar process for spike generation and information transformation. In the following sections, we explain the details of SNN architectures and learning methods applied to these types of networks. 

% Subsection 2.1
\subsection{SNN Architecture}
An SNN architecture consists of spiking neurons and interconnecting synapses that are modeled by adjustable scalar weights. The first step in implementing an SNN is to encode the analog input data into the spike trains using either a rate based method~\cite{gerstner2002spiking,gerstner2014neuronal}, some form of temporal coding~\cite{bohte2004evidence,hopfield1995pattern}, or population coding~\cite{bohte2002unsupervised}. As stated earlier, a biological neuron in the brain (and similarly in a simulated spiking neuron)  receives synaptic inputs form other neurons in the neural network. Biological neural networks have both action potential generation dynamics and network dynamics.
In comparison to true biological networks, the network dynamics of artificial SNNs are highly simplified.
In this context, it is useful to assume that the modeled spiking neurons have pure threshold dynamics (in contrast to, e.g., refractoriness, 
hysteresis, resonance dynamics, or post-inhibitory rebound properties). 
The activity of pre-synaptic neurons modulates the membrane potential of postsynaptic neurons, generating an action potential or spike when the membrane potential crosses a threshold. 
%The spike propagates along to axon which repeatedly branches to make contact with a large number of postsynaptic neurons.
Hodgkin and Huxley were the first to model this phenomenon \cite{hodgkin1952quantitative}.
Specifically, they created a model of action potential generation from the voltage gating properties of
the ion channels in the squid cell membrane of the squid axon.
%From this model, they were able to predict with high accuracy the speed at which the action potential 
%traveled down the giant axon of the squid.
After the Hodgkin and Huxley model with extensive biological details and high computational cost~\cite{hodgkin1952quantitative,gerstner2002spiking,kistler1997reduction}, diverse neuron models have been proposed such as the spike response model (SRM)~\cite{jolivet2003spike}, the Izhikevich neuron model~\cite{izhikevich2003simple}, and the leaky integrated-and-fire (LIF) neuron~\cite{delorme1999spikenet}. 
The LIF model is extremely popular because it captures the intuitive properties of external input accumulating charge across a leaky cell membrane with a clear threshold.
%For the SRM model, an attractive feature is that it easily allows for an adaptive threshold. 

Spike trains in a network of spiking neurons are propagated through synaptic connections.
A synapse can be either excitatory, which increases the neuron's membrane potential upon receiving input, or inhibitory, which decreases the neuron's membrane potential~\cite{kandel2000principles}. %,dale1935pharmacology,eccles1954cholinergic,strata1999dale}. 
The strength of the adaptive synapses (weights) can be changed as a result of learning. The learning rule of an SNN is its most challenging component for developing multi-layer (deep) SNNs, because the non-differentiability of spike trains limits the popular backpropagation algorithm.

% Subsection 2.2
\subsection{Learning Rules in SNNs}

% Deleted

% Added
As previously mentioned, in virtually all ANNs, spiking or non-spiking, learning is realized by adjusting scalar-valued synaptic weights.
Spiking enables a type of bio-plausible learning rule that cannot be directly replicated in non-spiking networks.
Neuroscientists have identified many variants of this learning rule that falls under the umbrella term
spike-timing-dependent plasticity (STDP).
Its key feature is that the weight (synaptic efficacy) connecting a pre- and post-synaptic neuron is adjusted according to their
relative spike times within an interval of roughly tens of milliseconds in length \cite{caporale2008spike}.
The information used to perform the weight adjustment is both local to the synapse and local in time.
The following subsections describe common learning mechanisms in SNNs, both unsupervised and supervised.

% Sussubsection 2.2.1
\subsubsection{Unsupervised Learning via STDP}

% Deleted

% Added
As stated above, unsupervised learning in SNNs often involves STDP as part of the learning mechanism \cite{caporale2008spike,markram2011history}.
The most common form of biological STDP has a very intuitive interpretation.
If a presynaptic neuron fires briefly (e.g., $\approx$ 10 ms) before the postsynaptic neuron, the weight connecting them is strengthened.
If the presynaptic neuron fires briefly after the postsynaptic neuron, then the causal relationship between the temporal events
is spurious and the weight is weakened.
Strengthening is called long-term potentiation (LTP) and weakening is called long-term depression (LTD).
The phrase ``long-term'' is used to distinguish between very transient effects on the scale of a few 
ms that are observed in experiments.

Formula~\ref{eq:stdp1} below idealizes the most common experimentally observed STDP rule for a single pair of spikes
obtained by fitting to experimental data \cite{dan2006spike}.
\begin{equation}
\Delta w=
\begin{cases}
Ae^{\frac{-(|t_{pre}-t_{post}|)}{\tau}} & t_{pre}-t_{post}\leq 0 \ , \ \ A>0\\
Be^{\frac{-(|t_{pre}-t_{post}|)}{\tau}} & t_{pre}-t_{post}>0 \ , \ \ B<0
\end{cases}
\label{eq:stdp1} 
\end{equation}
$w$ is the synaptic weight. 
$A>0$ and $B<0$ are usually constant parameters indicating learning rates.
$\tau$ is the time constant (e.g., 15 ms) for the temporal learning window. 
The first of the above cases describes LTP while the second describes LTD. 
The strength of the effect is modulated by a decaying exponential whose magnitude is controlled by the time-constant-scaled time difference between the pre- and postsynaptic spikes.
Rarely,  do artificial SNNs use this exact rule.
They usually use a variant, either to achieve more simplicity or to satisfy a convenient mathematical property.

Besides the temporally and spatially local weight change described in Eq.~\ref{eq:stdp1}, STDP has known important temporally accumulated network-level effects. 
For instance, STDP affects a neuron's behavior in response to repeated spike patterns embedded in a possibly stochastic spike train. 
A neuron (equipped with STDP-trained synapses) in coincidence with similar volleys of spikes is able to concentrate on afferents that consistently fire early (shorter latencies)~\cite{song2000competitive,guyonneau2005neurons}. 
Spike trains in many areas of the brain are highly reproducible. 
Guyonneau et al.~\cite{guyonneau2005neurons} have shown that presenting repeated inputs to an SNN equipped with STDP shapes neuronal selectivity to the stimulus patterns within the SNN.
Specifically, they showed that the response latency of the postsynaptic potential is decreased as STDP proceeds. Reducing the postsynaptic latency results in faster neural processing. Thus, the neuron responds faster to a specific input pattern than to any other. In fact, the STDP rule focuses on the first spikes of the input pattern which contain most of the information needed for pattern recognition.
It has been shown that repeating spatio-temporal patterns can be detected and learned by a single neuron based on STDP~\cite{masquelier2008spike,masquelier2018optimal}. STDP can also solve difficult computational problems in localizing a repeating spatio-temporal spike pattern and enabling some forms of temporal coding, even if an explicit time reference is missing~\cite{masquelier2008spike,masquelier2009competitive}. Using this approach, more complex networks with multiple output neurons have been developed~\cite{masquelier2007,masquelier2010,tavanaei2016b,kheradpisheh2016a}.

% Subsubsection
%\subsubsection{Probabilistic Approaches Toward STDP}
\subsubsection{Probabilistic Characterization of Unsupervised STDP}
%A number of previous empirical 
Many studies have provided evidence that at least an approximate Bayesian analysis of sensory stimuli occurs in the brain~\cite{rao2002,doya2007,mozer2008,kording2004}. 
In Bayesian inference, hidden causes 
(such as presence of an object of a particular category)
are inferred using both prior knowledge and the likelihood of new observations to obtain a posterior probability of the possible cause. 
%The question is 
Researchers have considered the possible role of probabilistic (Bayesian) computation as a primary information processing step in the brain in terms of STDP.

Nessler et al. (2009)~\cite{nessler2009} showed that a form of STDP, when used with Poisson spiking input neurons coupled with the appropriate stochastic 
winner-take-all (WTA)
circuit, is able to approximate a stochastic online expectation maximization (EM) algorithm to learn the parameters for a multinomial mixture distribution.
The model was intended to have some biological plausibility. 
%The architecture consists of a feedforward layer connected to a winner-take-all (WTA) circuit of $K$ stochastic spiking output neurons, $z_k$, for modeling the input data. 
The STDP rule used in their network is shown in Eq.~\ref{nessler}. 
LTP occurs if the presynaptic neuron fires briefly (e.g., within $\epsilon=10$ ms) before the postsynaptic neuron.
Otherwise LTD occurs. 
Generating a spike by an output neuron creates a sample from the coded posterior distribution of hidden variables which can be considered as the E-step in the EM algorithm. The application of STDP to the synapses of fired output neurons specifies the M-step in EM\@.
Nessler et al.\ (2013)~\cite{nessler2013} extended their network by using an inhibitory neuron to implement the WTA in order to improve the compatibility of the model for embedding in a cortical microcircuit.

\begin{equation}
\Delta w_{ki}=
\begin{cases}
&e^{-w_{ki}}-1, \ \ 0< t_{k}^\mathrm{f} - t_{i}^\mathrm{f} < \epsilon\\
&-1, \ \ \ \ \ \ \ \ \ \ \mathrm{otherwise}\\
\end{cases}
\label{nessler}
\end{equation}

Building on the stochastic WTA circuits described above, Klampfl and Maass (2013)~\cite{klampfl2013} developed a liquid state machine (LSM) containing input neurons, a reservoir of the WTA circuits, and a linear output readout. Further extension showed that STDP, applied on both the lateral excitatory synapses and synapses from afferent neurons, is able to represent the underlying statistical structure of such spatio-temporal input patterns~\cite{kappel2014}. In this framework, each spike train generated by the WTA circuits can be viewed as a sample from the state space of a hidden Markov model (HMM).

One drawback of the STDP model introduced in~\cite{nessler2009,nessler2013} is that its excitatory synaptic weights are negative. 
This, however, can be solved by shifting the weights to a positive value by using a constant parameter in the LTP rule. 
Based on this idea, Tavanaei and Maida~\cite{tavanaei2015a,tavanaei2016a} proposed an unsupervised rule for spatio-temporal pattern recognition and spoken word classification. It has been shown that the EM acquired in an SNN is able to approximately implement the EM algorithm in a Gaussian mixture model (GMM) embedded in the HMM states~\cite{tavanaei2016a}.

Using probabilistic learning in spiking neurons for modeling hidden causes has recently attracted attention. 
Rezende et al. (2011)~\cite{rezende2011} developed a bio-plausible learning rule based on the joint distribution of perceptions and hidden causes to adapt spontaneous spike sequences to match the empirical distribution of actual spike sequences~\cite{rezende2011}. 
The learning strategy involved minimizing the Kullback-Leibler divergence~\cite{kullback1951} as a non-commutative distance measure between the distribution representing the model (SNN) and a target distribution (observation). The EM algorithm in recurrent SNNs~\cite{brea2011} and probabilistic association between neurons generated by STDP in combination with intrinsic plasticity~\cite{pecevski2016} are two other instances of probabilistic learning in SNNs. The probabilistic rules also have been employed in sequential data processing~\cite{zemel2004} and Markov chain Monte Carlo sampling interpreted by stochastic firing activity of spiking neurons~\cite{buesing2011}.

% Subsubsection
\subsubsection{Supervised Learning}
\label{subsubSupervisedLearning}

%\sout{Although supervised learning has a long history in machine learning and neural networks, here we only review the recent studies that have developed supervised learning in SNNs. 
%In an SNN, the output neurons or their spike times have specific labels according to the target pattern category. 
%Supervised learning is based on minimizing the error between desired and output spike trains (also known as cost function) after receiving input stimuli.}

All supervised learning uses labels of some kind.
Most commonly, supervised learning adjusts weights via gradient descent on a cost function comparing observed and desired network outputs.
In the context of SNNs, supervised learning tries to minimize the error between desired and output spike trains, sometimes called readout error,
in response to inputs.

\paragraph{SNN Issues in Relation to Backpropagation}
From a biological vantage point, there has been considerable skepticism about whether the backpropagation training
procedure can be directly implemented in the brain.
With respect to SNNs, there are two prominent issues which can be seen from the formula below.
Shown below is a core formula, obtained from the chain rule, that occurs in all variants of backpropagation
\cite{Bishop1995a}.

\begin{equation}
\label{EqBackpropCore}
\delta_j^\mu = g^\prime(a_j^\mu) \sum_{k} w_{kj} \delta_k^\mu
\end{equation}

\noindent
In the above, $\delta_j^\mu$ and $\delta_k^\mu$ denote the partial derivative of the cost function
for input pattern $\mu$ with respect to the net input to some arbitrary unit $j$ or $k$.
Unit $j$ projects direct feedforward connections to the set of units indexed by $k$.
$g(\cdot)$ is the activation function applied to the net input of unit $j$, where that net input
is denoted $a_j^\mu$.
$w_{kj}$ are the feedforward weights projecting from unit $j$ to the set of units indexed by $k$.

Both parts of the RHS of Eq.~\ref{EqBackpropCore} present complications for bio-plausible spiking versions of
backpropagation.
First, the expression $g^\prime(\cdot)$ requires $g(\cdot)$ with respect to $w_{kj}$.
Since $g(\cdot)$ applies to a spiking neuron, it is likely represented by a sum of Dirac delta functions, which
means the derivative does not exist.
The second, and more serious complication, applies to both spiking and non-spiking networks
and was apparently first pointed out by Grossberg \cite[p.\ 49]{Grossberg1987a} and termed the ``weight transport'' problem.
The problem is the following.
The expression $\sum_{k} w_{kj}\delta_k^\mu$ is using the feedforward weights $w_{kj}$ in a feedback fashion.
This means that matching symmetric feedback weights must exist and project accurately to the correct neurons
(point-to-point feedback) in order for Eq.~\ref{EqBackpropCore} to be useable.

In the literature, the first issue has generally been addressed by using substitute or approximate derivatives.
One must be aware that some of these solutions are not bio-plausible.
For example, using the membrane potential of the presynaptic neuron as a surrogate becomes problematic because
its value is not local to the synapse (cf.
\cite{lee2016training}, 
Section~\ref{subsecDeepFullyConnected} of this review).
These approaches, however, are still useful from both engineering and scientific standpoints.

Progress on the second issue has recently been made by
\cite{Lillicrap2016a} and \cite{Zenke2018a}.
It was shown in \cite{Lillicrap2016a} that for some tasks, backpropagation could still perform well
if random feedback weights were used.
The authors in \cite{Zenke2018a} 
explored this further, examining three kinds of feedback (uniform,
random, and symmetric).
They found that simpler problems could be solved by any kind of feedback whereas complex problems needed
symmetric feedback.

\paragraph{Some Supervised Learning Methods for SNNs}

SpikeProp \cite{bohte2002} appears to be the first algorithm to train SNNs by backpropagating errors.
Their cost function took into account spike timing and SpikeProp was able to classify non-linearly separable data
for a temporally encoded XOR problem using a 3-layer architecture.
One of their key design choices was to use Gerstner's \cite{gerstner2002spiking} 
spike-response model (SRM) for the spiking neurons.
Using the SRM model, the issue of taking derivatives on the output spikes of the hidden units was avoided because those units'
responses could be directly modeled as continuous-valued PSPs applying to the output synapses that they projected to.
One limitation of this work is that each output unit was constrained to discharge exactly one spike.
Also, continuous variable values, such as in the temporally extended XOR problem, had to be encoded as spike-time
delays which could be quite long.

%\textbf{English not understandable in para below. Dates of research incompatible with improvements on SpikeProp.
%Revise or delete.}
%\sout{
%Further extensions of the SpikeProp algorithm resulted to RProp and QuickProp that had faster learning~\cite{mckennoch2006}.
%RProp is a method in training the learning rate based on the sign of the gradient instead of its magnitude~\cite{reed1998}.
%QuickProp approximates the global error surface (error gradient) by evaluating the local one using Newton's method}~\cite{reed1998,fahlman1988}.
%\sout{The initial implementation of the SpikeProp algorithm was used to classify the patterns coded in single spikes.}

Later advanced versions of SpikeProp, Multi-SpikeProp, were applicable in multiple spike coding~\cite{booij2005,ghosh2009}. 
Using the same neural architecture of SpikeProp, new formulations of temporal spike coding and spike time errors have recently improved the spiking backpropagation algorithm~\cite{liu2017mt,mostafa2017supervised}. 
The most recent implementation of backpropagation in SNNs has been proposed by Wu et. al. (2017)~\cite{wu2017spatio} who developed spatio-temporal gradient descent in multi-layer SNNs. 

More recent approaches to supervised training of SNNs include ReSuMe (remote supervised learning) \cite{ponulak2010,Kasinski2006a},
Chronotron \cite{florian2012}, and SPAN (spike pattern association neuron) \cite{mohemmed2012,mohemmed2013},
among others.
All of the above models consist of a single spiking neuron receiving inputs from many spiking presynaptic neurons.
The goal is to train the synapses to cause the post-synaptic neuron to generate a spike train with desired spike times.

ReSuMe adapts the Widrow-Hoff (Delta) rule, originally used for non-spiking linear units, to SNNs.
The Widrow-Hoff rule weight changes is proportional to the desired output minus the observed output, as shown below.

\begin{equation}
\Delta w = (y^\mathrm{d}- y^\mathrm{o})x = y^\mathrm{d}x-y^\mathrm{o}x
\end{equation}

\noindent
where $x$ is the presynaptic input and $y^\mathrm{d}$ and $y^\mathrm{o}$ 
are the desired and observed outputs, respectively.
When expanded as shown on the RHS and reformulated for SNNs,
the rule can be expressed as a sum of STDP and anti-STDP\@.
That is, the rule for training excitatory synapses takes the form

\begin{equation}
\Delta w = \Delta w^\mathrm{STDP}(S^\mathrm{in}, S^\mathrm{d}) + \Delta w^\mathrm{aSTDP}(S^\mathrm{in}, S^\mathrm{o}) .
\end{equation}

\noindent
In the above, $\Delta w^\mathrm{STDP}$ is a function of the correlation of the presynaptic and desired spike trains,
whereas $\Delta w^\mathrm{aSTDP}$ 
depends on the presynaptic and observed spike trains.
Because the learning rule uses the correlation between the teacher neuron (desired output) and the input neuron,
there is not a direct physical connection. 
This is why the word ``remote'' is used in the phrase ``remote supervised learning.''
Although it is not apparent in the above equation, the learning is constrained to fall with typical STDP eligibility windows.
The Chronotron was developed to improve on the Tempotron
\cite{gutig2006} 
which had the ability to train single neurons to recognize
encodings by the precise timing of incoming spikes.
The limitation of the Tempotron was that was restricted to outputting 0 or 1 spikes during a predetermined interval.
Because of this, the output did not encode spike timing information.
This precluded the ability of a Tempotron to meaningfully send its output to another Tempotron.
The motivation of the Chronotron was similar to that of SpikeProp and its successors.
The innovation of the Chronotron was to base the supervised training on a more sophisticated distance measure,
namely the Victor-Purpora (VP) distance metric \cite{victor1997} 
between two spike trains.
This metric is ``the minimum cost of transforming one spike train into the other by creating, removing, or moving spikes.'' \cite[p.\ 3]{florian2012}
They adapted the VP distance so that it would be piecewise differentiable and admissible as a cost function to perform gradient descent
with respect to the weights.

Similar to ReSuMe, the SPAN model develops its learning algorithm from the Widrow-Hoff rule.
However, instead of adapting the rule to an SNN, SPAN makes the SNN compatible with Widrow-Hoff
by digital-to-analog conversion of spike trains using alpha kernels of the form
$te^\frac{-t}{\tau}$.
As this is a common formula for modeling a postsynaptic potential, this step in effect converts all spikes to a linear
summation of PSPs.
Note that this is similar to the technique that was used in SpikeProp described at the beginning of this subsection.
The learning rule can then be written as

\begin{equation}
\Delta w \propto \int \tilde{x}_i (\tilde{y}_\mathrm{d}(t) - \tilde{y}_o(t)) dt
\end{equation}

\noindent
where the tilde symbol indicates the analog version of the spike train and the bounds of integration
cover the relevant local time interval.

In \cite{huh2017gradient}, 
it was observed that the previous gradient-based learning methods all still had the
constraint that the number and times of output spikes must be prespecified which placed limits on their applicability.
They replaced the hard spike threshold with a narrow support gate function, $g(\cdot)$,
such that $g(v) \ge 0$ and $\int g dv = 1$,
and $v$ 
is the membrane potential.
Intuitively, this allows modeled postsynaptic currents to be released when the membrane potential approaches
threshold leading to continuity in the spike generation mechanism.
Experimentally,
it was found that weight updates occured near spike times ``bearing close resemblence to reward-modulated STDP'' \cite[p.\ 8]{huh2017gradient},
which enhances the biological relevance of the model.

In \cite{tavanaei2017bp}, a supervised learning method was proposed (BP-STDP) where the backpropagation update rules 
were converted to temporally local STDP rules for multi-layer SNNs. 
This model achieved accuracies comparable to equal-sized conventional and spiking networks for the MNIST benchmark (see Section~\ref{subsecDeepFullyConnected}). 

Another implementation of supervised learning in SNNs can be based on optimizing the likelihood and probability of the postsynaptic spikes to match the desired ones. Pfister et al. (2006)~\cite{pfister2006} developed a model to optimize the likelihood of postsynaptic firing at one or several desired times. They proposed a modified version of the SRM neuronal model such that it uses a stochastic threshold on the membrane potential.

In another approach %, the activity of the output neuron is considered as the objective function in 
to supervised learning,
%That is, 
each output neuron represents a class of data (pattern). Output neurons are in competition to be selected and responsive to the input patterns. In this approach, firing the target neuron causes STDP over the incoming synapses and firing the non-target neurons causes anti-STDP. 
This approach has successfully been used in SNNs for numerical data classification~\cite{wang2014}, handwritten digit recognition~\cite{tavanaei2015b}, spoken digit classification~\cite{tavanaei2016b}, and reinforcement learning in SNNs~\cite{mozafari2017first}. The sharp synaptic weight adaptation based on immediate STDP and anti-STDP results in fast learning.

\section{Deep Learning in SNNs}
\begin{figure}
\centering
\includegraphics[scale=.55]{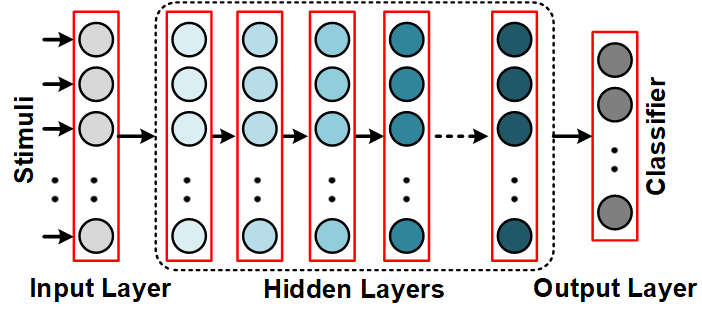}
\caption{Simplest deep neural architecture, usually fully connected, with input, hidden, and output layers. 
The input layer learns to perform pre-processing on the input. 
The information is then sent to a series of hidden layers, the number of which can vary. 
As the information propagates through hidden layers, more complex features are extracted and learned. 
The output layer performs classification and determines the label of the input stimulus, usually by softmax (see text). }
\label{fig:dnn}
\end{figure}

Deep learning uses an architecture with many layers of trainable parameters and has demonstrated outstanding performance in machine learning and AI applications~\cite{lecun2015deep,schmidhuber2015}. Deep neural networks (DNNs) are trained end-to-end by using optimization algorithms usually based on backpropagation. The multi-layer neural architecture in the primate's brain has inspired researchers to concentrate on the depth of non-linear neural layers instead of using shallow networks with many neurons. 
Also, theoretical and experimental results show better performance of deep
rather than wide  structures~\cite{goodfellow2016deep,kheradpisheh2016deep,kheradpisheh2016humans}. 
Deep neural networks extract complex features through sequential layers of neurons equipped by non-linear, differentiable activation functions to provide an appropriate platform for the backpropagation algorithm. Fig.~\ref{fig:dnn} depicts a deep NN architecture with several hidden layers.

For most classification problems, the output layer of a deep network uses a 
softmax module.
The training vectors use a one-hot encoding.
In a one-hot encoding each vector component corresponds to one of the possible classes.
This vector is binary 
with exactly one component set to 1 that corresponds to the desired target class.
The softmax module for the output layer guarantees that the values of each of the output units falls within the range (0, 1) and also sum to 1.
This gives a set of mutually exclusive and exhaustive probability values.
The softmax formula, sometimes called the normalized exponential, is given below

\begin{equation}
y_i = \frac{\exp(a_i)}{\sum_j \exp(a_j)}
\end{equation}

\noindent
where, $a_i$, 
is the net input to a particular output unit, $j$ 
indexes the set of output units, and $y_i$ 
is the value of output unit $i$, 
which
falls in the range (0, 1).

In addition to the fully connected architecture in Fig.~\ref{fig:dnn} 
and discussed in Section~\ref{subsecDeepFullyConnected}, 
there are also deep convolutional neural networks (DCNNs) discussed in Section~\ref{subsecDeepSpikingCNNs}, 
deep belief networks (DBNs) discussed in Section~\ref{subsecSpikingDBNs}, and recurrent neural networks (RNNs) discussed in Section~\ref{subsecrnn}.

%A special deep learning architecture known as a deep convolutional neural network (DCNN) contains stacked layers of convolution and pooling followed by one or more layers of a fully connected network. DCNNs have shown very good performance in image recognition~\cite{krizhevsky2012,szegedy2015going}, speech recognition~\cite{sainath2013deep}, healthcare and bioinformatics~\cite{tavanaei2016towards}. 
%Another successful approach in deep learning is the deep belief network (DBN)~\cite{bengio2009learning,hinton2006} containing stacked restricted Boltzmann machines (RBM) for unsupervised feature extraction.

SNNs have also shown promising performance in a number of pattern recognition tasks~\cite{ghosh2009b,kasabov2014neucube}. 
However, the performance of directly trained spiking deep networks are not as good as traditional DNNs represented in the literature. 
Therefore, a spiking deep network (spiking DNN, spiking CNN,spiking RNN, or spiking DBN) with good performance comparable with traditional deep learning methods, 
is a challenging topic because of its importance in DNN hardware implementations.

Masquelier and Thorpe (2007, 2010) developed one of the earliest feedforward hierarchical convolutional network of spiking neurons for unsupervised learning of visual features~\cite{masquelier2007,masquelier2010}. This network was extended for larger problems, such as~\cite{kheradpisheh2016a}. 
Using an STDP rule with a probabilistic interpretation, 
the performance of the model was later improved in different object recognition tasks~\cite{tavanaei2016c}.
Further attempts led to several multi-layer SNNs, with STDP learning, that performed greatly in adaptive multi-view pattern recognition~\cite{wysoski2008} and handwritten digit recognition~\cite{beyeler2013}. These models mostly used one or more layers for pre-processing, one learning layer, and one classifier (output neuron) layer. Although these networks are known as multi-layer SNNs, they do not offer multi-layer learning. Specifically, these SNNs are limited by using only one trainable layer, even though they have many layers of processing.

Encouraged by the power-efficiency and biological plausibility of neuromorphic platforms, 
a number of recent studies have concentrated on developing deep SNNs for these platforms. 
Previous studies exploring supervised and unsupervised learning rules in spiking architectures can be employed to develop hierarchies of feature extraction and classification modules. Existing deep SNNs do not perform as accurately as the traditional deep learning models. However, SNNs enable power-efficient platforms mimicking the brain functionality for solving complex problems, especially in new trends of autonomous objects. Therefore, developing a neural network that is as efficient and biologically plausible as SNNs but as powerful as DNNs in performing different tasks can be the next challenging topic in the field of artificial intelligence and computational neuroscience. 
The initial steps for implementing a deep spiking framework can be fulfilled by either converting the trained neural networks to a spiking platform or using spike-based, online learning methods. 

The remaining subsections review spiking deep learning approaches covering deep fully connected SNNs, spiking CNNs, spiking DBNs, and spiking RNNs.

% The proposed SDCNNs in the literature almost use offline learning using a traditional DCNN and then the trained synaptic weights are transferred to spiking neuron frameworks~\cite{diehl2015,cao2015,matsugu2002,garbin2014}. Similarly, O'Connor et al. (2013) introduced an approach to implement a spiking DBN by converting a pre-trained DBN to an SNN~\cite{o2015}. Neftci et al. (2014) developed an event-driven variation of Contrastive Divergence (CD) to train an RBM based on IF neurons~\cite{neftci2014}. However, 

% Subsection
\subsection{Deep, Fully Connected SNNs}
\label{subsecDeepFullyConnected}
Recent studies have developed a number of deep SNNs using STDP and stochastic gradient descent.
Spiking networks consisting of many LIF neurons equipped by spike-based synaptic plasticity rules have shown success in different pattern recognition tasks~\cite{brader2007learning,eliasmith2012large}. Diehl et al.~\cite{diehl2015unsupervised} showed that STDP in a two-layer SNN is able to extract discriminative features and patterns from stimuli. 
They used unsupervised learning rules introduced by~\cite{morrison2007spike,nessler2013,pfister2006triplets} to train the SNN for 
the Modified National Institute of Standards and Technology (MNIST) dataset~\cite{lecun1998mnist} digit recognition with the best performance of 95\%.    

\begin{figure}[h]
\centering
\includegraphics[scale=.75]{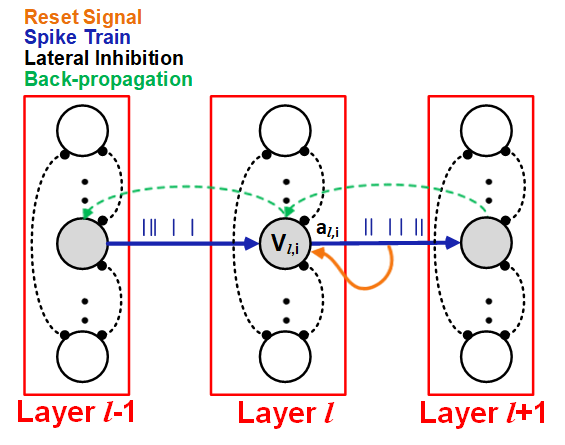}
\caption{Deep SNN equipped with backpropagation proposed by Lee et. al.~\cite{lee2016training}. The neuron's activation value, $a_{l,i}$, is given by the neuron's membrane potential. The differentiable activation function, which is calculated by the neuron's excitatory input, lateral inhibition, and threshold, is used for developing backpropagation using the chain rule. The output activation value of the current layer (layer $l$) is used as input for the next layer in the backpropagation algorithm.}
\label{fig:dsnnlee}
\end{figure}

Towards linking biologically plausible learning methods and conventional learning algorithms in neural networks, a number of deep SNNs have recently been developed. For example, Bengio et al.~\cite{bengio2015towards} proposed a deep learning method using forward and backward neural activity propagation. The learning rule in this network is based on the idea that STDP implements the gradient descent learning rule~\cite{hinton2007backpropagation,bengio2017stdp}. Using pre- and postsynaptic spike trains, O'Connor and Welling~\cite{o2016deep} developed a backpropagation algorithm in deep SNNs using the outer product of pre- and postsynaptic spike counts. 
They showed high performance of the spiking multi-layer perceptron on the MNIST benchmark (97.93\%) which is comparable to the performance of the conventional deep neural networks equipped with rectified linear units (ReLUs) of 98.37\%. 
Recently, Lee et al. (2016)~\cite{lee2016training} proposed a backpropagation algorithm by treating the neuron's membrane potential as the differentiable signal to act analogous to the non-linear activation functions in traditional neural networks (Fig.~\ref{fig:dsnnlee}). 
%In this model, the membrane potential, $V$, of an LIF neuron (neuron $i$ in the network of $n$ neurons) receiving $m$ presynaptic spike trains at time $t$ is computed by
%\begin{equation}
%\label{eq:dsnn1}
%V_i(t)=\sum_{k=1}^m w_{ik}x_k(t) - V_{\mathrm{th},i}a_i(t) +0.5 V_{\mathrm{th},i} \sum_{j=1, j\neq i}^n \gamma_{ij} a_j(t)
%\end{equation} 
%Where, $w_{ik}$'s are the synaptic weights, $V_{\mathrm{th},i}$ is the threshold, and $-1\leq \gamma_{ij} \leq 0$ denotes the strength of inhibition. The input activation, $x_k(t)$, and the output transfer function, $a_i(t)$, with respect to input spikes, $t_p$, and output spikes, $t_q$, are defined as follows
%\begin{equation}
%\label{eq:dsnn2}
%x_k(t) = \sum_p exp(\frac{t_p-t}{\tau})\ , \ \ \ \ a_i(t) = \sum_q exp(\frac{t_q-t}{\tau})
%\end{equation}   
%$\tau$ is the membrane time constant. The backpropagation equations are obtained by transfer functions of LIF neurons (where $V_i(t)=0$) as follows
%\begin{equation}
%\label{eq:dsnn3}
%a_i\approx \frac{\sum_{k=1}^m w_{ik}x_k}{V_{\mathrm{th},i}} + 0.5 \sum_{j=1, j\neq i}^n \gamma_{ij} a_j
%\end{equation}
The performance of 98.88\% on the MNIST dataset was reported in this study while the number of computational operations were five times fewer than traditional DNNs, in their experiments.
To further reduce the computational cost of learning in these deep SNNs, Neftci et. al. (2017)~\cite{neftci2017event} proposed an event-driven random backpropagation (eRBP) algorithm simplifying the backpropagation chain path. 
The eRBP rule used an error-modulated synaptic plasticity in which all the information used for learning was locally available at the neuron and synapse~\cite{neftci2017event}. 

\begin{figure}[h]
\centering
\includegraphics[scale=.8]{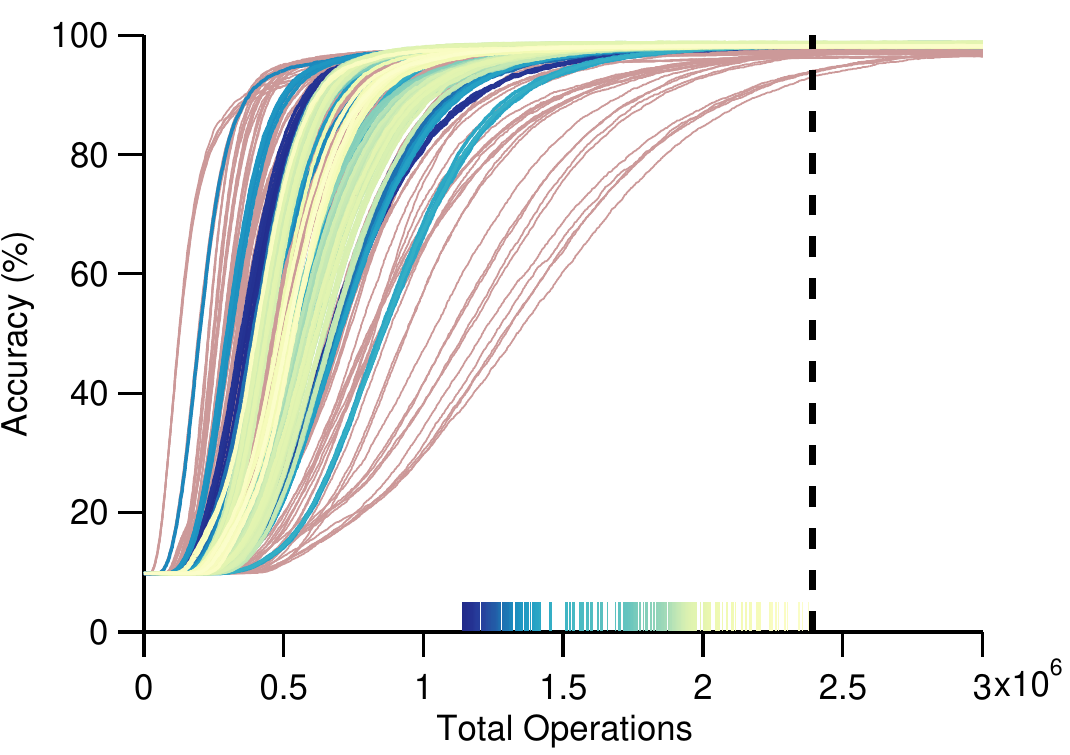}
\caption{The total number of operations needed to achieve a given accuracy for MNIST classification by deep SNNs converted from an offline trained deep neural network in comparison with the traditional (non-spiking) deep neural network~\cite{neil2016learning}. The vertical dashed line shows the number of operations required for the non-spiking deep neural network to achieve the accuracy of 98\%. The other curves show the accuracy of 522 deep SNNs (with different network setups) versus the number of operations. The pink curves show the networks that achieve less than 98\% accuracy within the computing constraint. The
colored vertical lines on the horizontal axis indicate the number of operations at which the corresponding SNNs reached 98\% accuracy.}
\label{fig:dsnnneil}
\end{figure}

A more direct approach to taking advantage of power-efficient SNNs is to convert an offline trained DNN to a neuromorphic spiking platform (ANN-to-SNN), specifically for hardware implementation~\cite{querlioz2013immunity}. 
To substitute for the floating-point activation values in DNNs, 
rate-based coding is generally used in which higher activations are replaced by higher spike rates. 
Using this approach, several models have been developed that obtained excellent accuracy
performance~\cite{diehl2015,esser2015backpropagation,rueckauer2017conversion,stromatias2017event}. 
In a separate effort
to assess the power consumption of deep SNNs, Neil et al~\cite{neil2016learning}
studied many different models, all of which achieved
the same accuracy rate of 98\% on the MNIST digit recognition task.
They all used the same 784-1200-1200-10 three-layer architecture but
applied optimized parameters and SNN architecture settings, in ANN-to-SNN conversion, to reduce the power and latency of the model. 
The performance of a DNN and its converted deep SNN versus the total required operations is shown in Fig.~\ref{fig:dsnnneil}.

% Subsubsection
\subsection{Spiking CNNs}
\label{subsecDeepSpikingCNNs}
\begin{figure*}
\centering
\includegraphics[scale=.47]{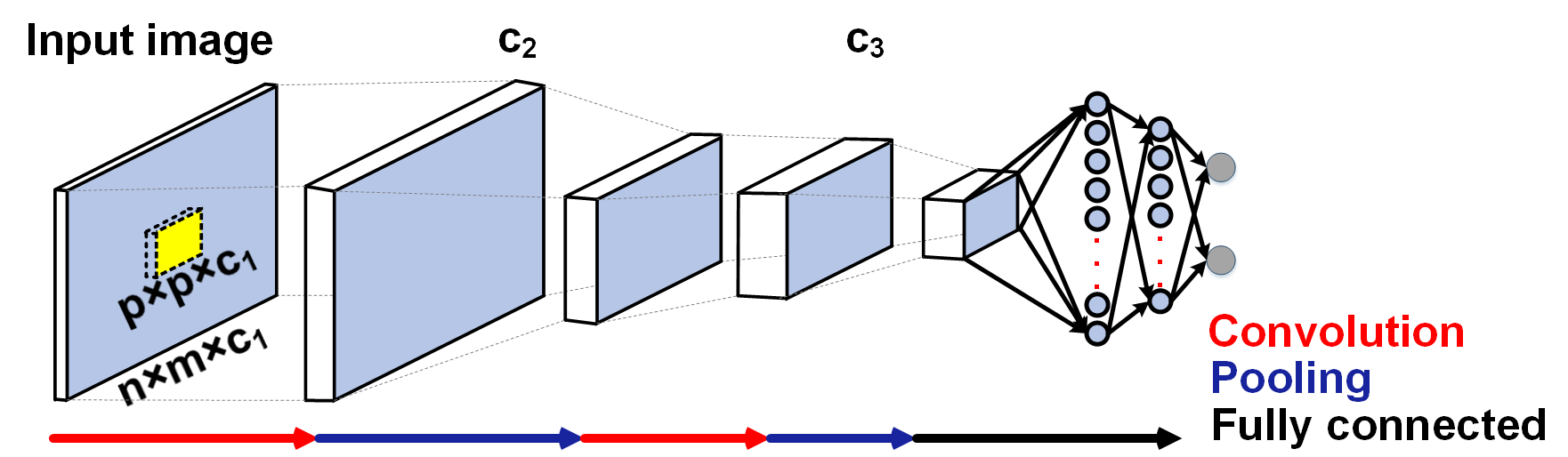}
\caption{LeNet: Early CNN proposed by LeCun et.~al.~\cite{lecun1998gradient,lecun2015lenet}. The network consists of two convolutional/pooling layers followed by 
fully connected layers for image classification. }
\label{fig:lenet}
\end{figure*}

Deep convolutional neural networks (DCNNs) are mostly used in applications involving images.
They consist of a sequence of convolution and pooling (sub-sampling) layers followed by a feedforward classifier like that in Fig.~\ref{fig:dnn}. 
This type of network has shown outstanding performance in image recognition~\cite{krizhevsky2012,rawat2017deep,oquab2014learning,simonyan2014very}, 
speech recognition~\cite{sainath2013deep,abdel2012applying,abdel2013exploring}, 
bioinformatics~\cite{zeng2016convolutional,quang2016danq,tavanaei2016towards}, object detection and segmentation~\cite{ronneberger2015u,long2015fully}, and so forth. 
Fig.~\ref{fig:lenet} shows the LeNet architecture, an early deep CNN, for image classification~\cite{lecun1998gradient,lecun2015lenet,szegedy2015going}. 
The question is how an spiking CNN with such an architecture can be trained while incorporating traditional CNN properties. 
In the case of vision, the first layer of convolution is interpreted as extracting primary visual features (
sometimes resembling oriented-edge detectors
modeled by the outputs of Gabor filters~\cite{marcelja1980mathematical}). 
Subsequent layers extract increasingly more complex features for classification purposes. 
The pooling layer performs subsampling and reduces the size of the previous layer using an arithmetic operation such as maximum or average over a 
square neighborhood of neurons in the relevant feature map. 
Later in the hierarchy, these layers develop invariance to changes in orientation, scale, and local translation.

The representational properties of early layers in the CNN mentioned above are similar to the response properties of neurons in primary visual cortex (V1), 
which is the first cortical area in the visual hierarchy of the primate's brain. 
For example, neurons in area V1 detect primary visual features, such as oriented edges, from input images~\cite{hubel1959receptive,hubel1962receptive}. 
Each V1 neuron is selective to a particular orientation, meaning that when a stimulus with this orientation is presented, only selective neurons to this orientation respond maximally. %Therefore, the presentation of a visual stimulus only drives spikes in a group of neurons that are selective to different attributes of input stimuli. 
Representation learning methods, which use neural networks such as autoencoders and sparse coding schemes, learn to discover visual features similar to the receptive field properties found in V1~\cite{foldiak1990forming,olshausen1996emergence,bell1997independent,rehn2007network}. Bio-inspired SNNs also have obvious footprints in representation learning using sparse coding~\cite{zylberberg2011sparse,king2013inhibitory}, independent component analysis (ICA)~\cite{savin2010independent}, and an STDP-based autoencoder~\cite{burbank2015mirrored}. 

\begin{figure*}
\centering
\includegraphics[scale=.55]{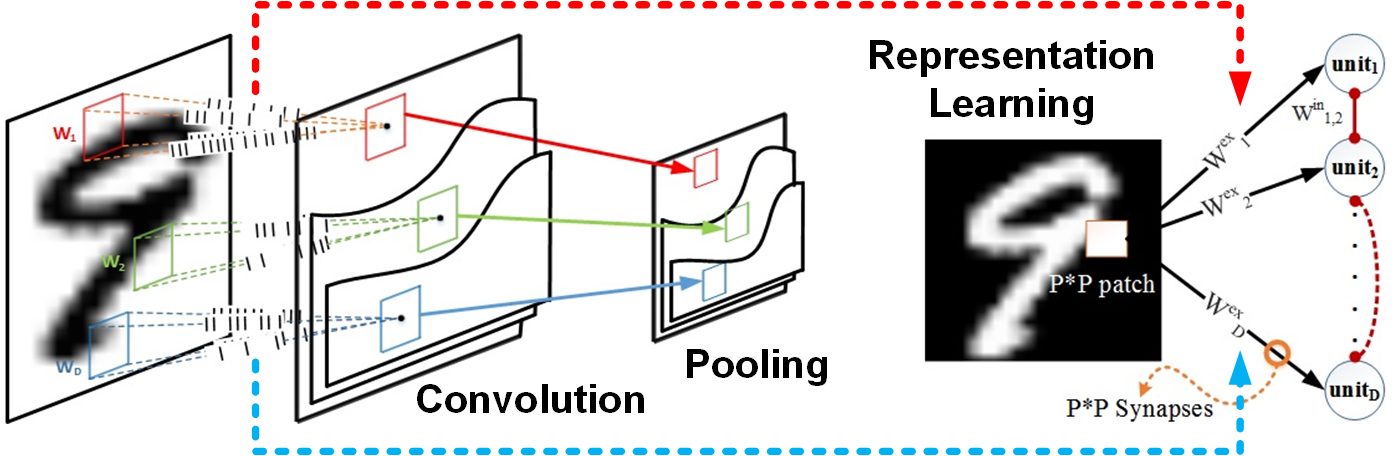}
\caption{Representation learning (SAILnet~\cite{zylberberg2011sparse}) for layer-wise unsupervised learning of a spiking CNN~\cite{tavanaei2017multi}. The excitatory synaptic weights connected to neurons in the representation layer specify convolutional filters. This architecture determines that representation learning in single-layer SNNs can be utilized to train layer-wise spiking CNNs. }
\label{fig:filterCNN}
\end{figure*}

As mentioned earlier, CNNs commonly use V1-like receptive field kernels in early layers to extract features from stimuli by convolving the kernels over the input (e.g. image). 
Subsequent layers combine the previous layer's kernels to learn increasingly complex and abstract stimulus features. 
Representation filters (trained or hand-crafted) and STDP learning rules can be used to develop spiking CNNs. A convolutional/pooling layer trained by a local spike-based representation learning algorithm is shown in Fig.~\ref{fig:filterCNN}.
Hand-crafted convolutional kernels have been used in the first layer of a number of spiking CNNs that have obtained high classification performance~\cite{masquelier2007,kheradpisheh2016a,tavanaei2016c,zhao2015feedforward,kheradpisheh2016stdp}.
Difference-of-Gaussian (DoG) is a common hand-crafted filter that is used to extract features in the early layers of SNNs. 
This choice is bio-motivated to mimic inputs to the mammalian primary visual cortex.
A recent study has used a layer of DoG filters as input layer of an SNN that is followed by more convolutional/pooling layers trained by STDP~\cite{kheradpisheh2016stdp}. 
This network architecture extracted visual features that were sent to an SVM classifier, yielding accuracy of 98.4\% on MNIST. 
To train convolutional filters, layer-wise spiking representation learning approaches have been implemented in recent spiking CNNs~\cite{tavanaei2016bio,tavanaei2017multi,panda2016unsupervised,tavanaei2018training}. Tavanaei et al. (2017)~\cite{tavanaei2016bio,tavanaei2017multi} used SAILnet~\cite{zylberberg2011sparse} to train orientation selective kernels used in the initial layer of a spiking CNN. The convolutional layer in this network is followed by a feature discovery layer 
equipped with an STDP variant~\cite{tavanaei2016c} to extract visual features for classification. 
Implementing the stacked convolutional autoencoders~\cite{panda2016unsupervised} showed further improvement in performance on MNIST (99.05\%), 
which is comparable to the traditional CNNs.
      
Non-spiking
CNNs are trained using the backpropagation algorithm. 
Recently, backpropagation has also been employed for training spiking CNNs~\cite{lee2016training,panda2016unsupervised}. 
Panda and Roy (2016)~\cite{panda2016unsupervised}, using approximations developed in~\cite{anwani2015normad}, 
showed how to build a hierarchical spiking convolutional autoencoder (AE) using backpropagation. 
The spiking convolutional autoencoder is an important module for enabling the construction of deep spiking CNNs. 
Their proof-of-concept implementation (SpikeCNN) used two learning layers on the MNIST dataset (handwritten digits)~\cite{lecun1998gradient} 
and three learning layers on the CIFAR-10 dataset (10-category tiny images)~\cite{krizhevsky2009learning}. 
They used local, layer-wise learning of convolutional layers while Lee et. al. (2016)~\cite{lee2016training} developed an end-to-end gradient descent learning method. 
Both methods used the neural membrane potential as replacements for differentiable activation functions to apply the backpropagation algorithm. 
Lee et.\ al.'s approach (for the spiking CNN )~\cite{lee2016training} showed better performance than the layer-wise convolutional autoencoders~\cite{panda2016unsupervised}. 
In these models, higher membrane potential correlates with higher spike probability.

\begin{figure*}
\centering
\includegraphics[scale=.55]{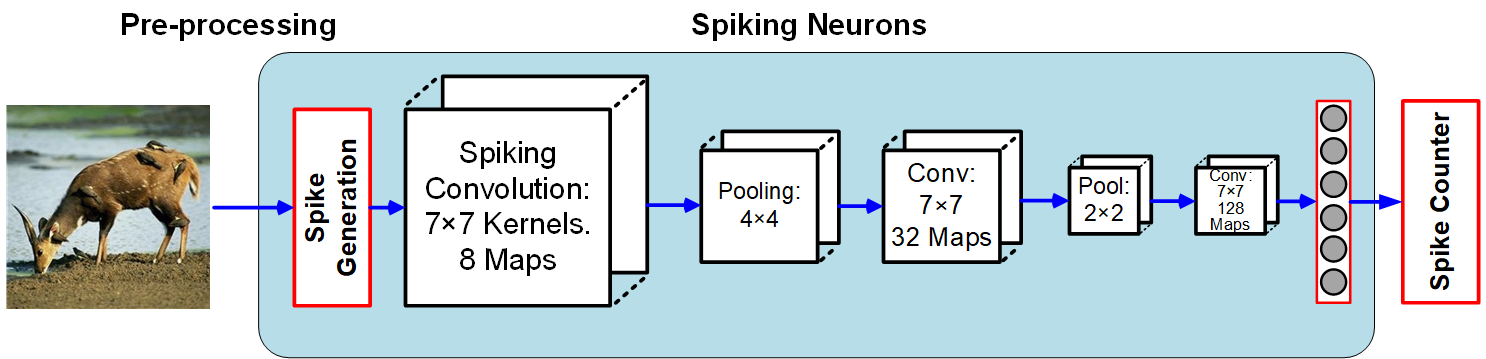}
\caption{Spiking CNN architecture developed by Cao et. al.~\cite{cao2015}. 
The input image, after pre-processing, is converted to spike trains based on the pixel intensity. 
The spiking  layers use the weights trained by a non-spiking CNN. 
The last component selects the neuron with maximum activity (spike frequency) as the image's class.}
\label{fig:cao}
\end{figure*}

The main approach to take advantage of spiking platforms while avoiding the training process of spiking CNNs is to convert an already trained CNN to a 
spiking architecture by using the trained synaptic weights, similar to the ANN-to-SNN conversion method. 
Many studies have shown high performance of converted spiking CNNs (close to conventional CNNs) while using fewer operations and consuming less energy~\cite{hunsberger2015spiking,hunsberger2016training,esser2015backpropagation,neil2016effective}, which enable the deep CNNs to be implemented on hardware~\cite{indiveri2015neuromorphic,garbin2014variability,esser2016convolutional}. 
One of the initial successful CNN-to-SNN conversion methods for energy efficient pattern recognition is the architecture shown in Fig.~\ref{fig:cao}~\cite{cao2015}. 
Later, Diehl et al. (2015)~\cite{diehl2015} improved this architecture using weight normalization to reduce performance loss. 
Recent work by Rueckauer et al.~\cite{rueckauer2016theory,rueckauer2017conversion} proposed several conversion criteria such that the new spiking CNN recognize's more difficult objects than MNIST (e.g. CIFAR-10 and ImageNet~\cite{deng2009imagenet}). 

\subsection{Spiking Deep Belief Networks}
\label{subsecSpikingDBNs}
\begin{figure}
\centering
\subfloat[]{
\includegraphics[scale=.55]{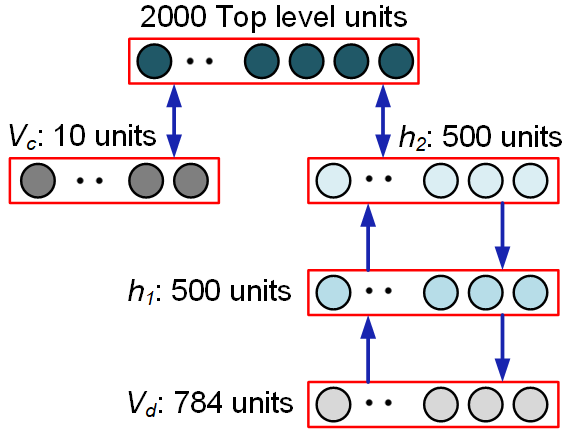}
\label{fig:dbn}
}\quad
\subfloat[]{
\includegraphics[scale=.5]{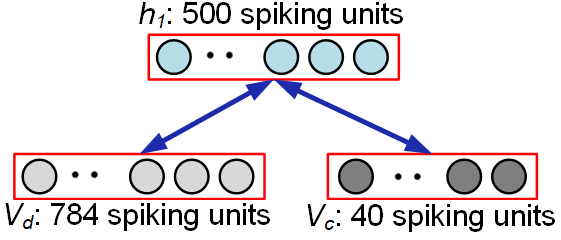}
\label{fig:dbn2}
}
\caption{(a): The DBN proposed by Hinton et. al.~\cite{hinton2006} for MNIST image classification. This network consists of three stacked RBMs with 500, 500, and 2000 representation neurons. The input and output include 784 (as the number of pixels, $28\times 28$) and 10 (as the number of classes, 0,...,9) neurons, respectively. (b): The spiking RBM architecture introduced by Neftci et. al.~\cite{neftci2014} consisting of 500 hidden neurons, 784 input neurons, and 40 class neurons (824 visible neurons).}

\end{figure}

Deep belief networks (DBNs) \cite{bengio2009learning} are a type of multi-layer network initially developed by Hinton et al. (2006)~\cite{hinton2006}. 
They efficiently use greedy layer-wise unsupervised learning and are made of stochastic binary units,
meaning that the binary state of the unit is updated using a probability function. 
The layer-wise method stacks pre-trained, single-layer learning modules known as restricted Boltzmann machines (RBMs). 
The representation layer in an RBM is restricted from having lateral connections.
This enables the learning algorithm to optimize the representation by making use of independence assumptions among the representation units,
given a particular input state.
The original DBN architecture was successfully 
trained on the MNIST dataset and is shown in Figure~\ref{fig:dbn}. The RBMs are trained in a layerwise fashion by contrastive divergence (CD), which approximates a maximum-likelihood learning algorithm. 
Unlike backpropagation, the CD update equations do not use derivatives.
The pre-trained hierarchy is fine-tuned by backpropagation if labeled data are available. DBNs provide a layer-wise structure for feature extraction, representation, and universal approximation~\cite{salama2010deep,le2008representational,le2010deep}. 

Lee et. al. (2008)~\cite{lee2008sparse} used interleaving CD with gradient descent on the sparsity term to implement sparse DBNs which were used to model cortical visual areas V1 and V2. Further extensions of the model led to a convolutional sparse DBNs~\cite{lee2011unsupervised}. This was accomplished by redefining the energy function to be consistent with the tied weights of a convolutional network and then using Gibbs sampling to realize the appropriate weight update rules. DBNs and convolutional DBNs have successfully been employed in many areas such as visual processing~\cite{krizhevsky2010convolutional,susskind2008generating,mleczko2015rough,liu2014facial}, audio processing~\cite{lee2009unsupervised,kang2013multi,mohamed2009deep,hamel2010learning,mohamed2012acoustic}, time series forecasting~\cite{kuremoto2014time},
and protein folding \cite{jo2015improving}. 

The first step in developing a spiking DBN is to start with a spiking RBM.
Figure~\ref{fig:dbn2} shows the architecture of a spiking RBM introduced by
Neftci et al.\ (2014)~\cite{neftci2014}. A spiking RBM uses stochastic integrate-and-fire neurons instead of the memoryless stochastic units in a standard RBM. 
Neftci et al.\ (2014) showed that, in the context of a particular type of spiking network, a variant of STDP can approximate CD. That is, the learned distributions for spiking networks capture the same statistical properties as the learned distributions for the equivalent non-spiking networks, establishing an important foundational result. 
%They achieved 91.9\% performance on the MNIST image classification task. 

One approach toward developing functional spiking DBNs is to convert previously trained DBNs to spiking platforms similar to the conversion method explained 
for SNNs or spiking CNNs. 
The first spiking DBN was introduced by O'Connor et al. (2013)~\cite{o2013real} in which a DBN is converted to a network of LIF spiking neurons for MNIST image classification. 
This work was then extended~\cite{stromatias2015robustness} to develop a noise robust spiking DBN and as well as conforming to hardware constraints. 
The spiking DBNs and RBMs are power-efficient and this enables them to be implemented on low latency hardware with high accuracy close to that of traditional DBNs~\cite{stromatias2015scalable,merolla2011digital,neil2014minitaur}. 

Recently, it has been shown there is an equivalence between so called `hybrid' Boltzmann machines (HBMs)
\cite{barra2012equivalence} 
and Hopfield networks \cite{hopfield1982neural,Hertz1991a}.
An HBM is a restricted Boltzmann machine in which the hidden (representation) units can take on continuous values
(while the visible units still have binary values) \cite{barra2012equivalence}.
When the functions within the HBM are marginalized over the hidden units, both the Hopfield and the HBM systems have been shown to be
thermodynamically equivalent.
Although
Hopfield networks are primarily used as models of pattern association, the thermodynamic equivalence 
allows them to be simulated by HBMs.
In the HBM, the $N$ binary visible units correspond to binary stochastic neurons in the Hopfield network
and the $P$ hidden units in the HBM correspond to stored patterns in the Hopfield network.
HBMs offer a new way to simulate Hopfield networks while using fewer synapses.
Specifically,
a Hopfield network requires updating $N$ neurons and $N/(N-1)/2$ synapses, whereas the HBM requires $H+P$ neurons
and updating $HP$ synapses, where $P$ is the number of stored patterns.
Further developments of this theory are found in \cite{barra2017phase,barra2018phase,tubiana2017emergence}.

% Subsection
\subsection{Recurrent SNNs}
\label{subsecrnn}
A neural network is recurrent if its directed graph representation has a cycle.
Any network that has a winner(s)-take-all (WTA) module or a softmax module is at least implicitly recurrent
because equivalent function is implemented in the brain by mutually recurrent inhibitory connections.
Any network trained by backpropagation is also implicitly recurrent because the training algorithm (as explained in Section \ref{subsubSupervisedLearning})
presupposes the existence of recurrent connections.
Sections \ref{subsubGatedSNNs} and \ref{LSMReservoir} 
discuss gated SNNs and reservoir models, respectively.
Both types of models are intended to process sequential data.
The former processes spatiotemporal data and is usually intended as a model of cortical microcircuits.
The latter focuses more on sequential data only.

% Subsubsection
% Subsubsection
\subsubsection{Gated SNNs}
\label{subsubGatedSNNs}
Recurrent neural networks (RNNs) are used for processing temporal information.
The most common method to train RNNs is backpropagation through time (BPTT), which unrolls the recurrent
network for some number of steps into the past, and then trains the unrolled network as if it was a feedforward
network.
Since the same recurrent weights are shared through all unrolled layers for the resulting feedforward network,
there are issues in training for long sequences, specifically the emergence of vanishing and exploding gradients.
In the former case, the network stops learning and in the latter, the training becomes unstable \cite{bengio1994learning}.

Because of these problems, the research in \cite{hochreiter1997long}
introduced an innovation into recurrent networks called a constant error carousel (CEC) that avoided
repeated multiplication of derivatives.
%in training conventional RNNs that relate to vanishing and exploding gradients} \cite{bengio1994learning},
These have become known as gated recurrent networks and have virtually replaced traditional RNNs.
The first gated recurrent network was the long short-term memory (LSTM) \cite{hochreiter1997long}.
LSTMs and other gated recurrent networks (GRUs) \cite{chung2014empirical} 
are conventional ANNs in the sense that they do not use spiking neurons,
but they are also unconventional in the sense that they replace units having recurrent connections
with `cells' that contain state as well as gates and, because of this, can readily adapt to the structure of input sequences.
The gates control information flow into, out of, and
within the cells. 
The gates are controlled by trainable weights.
The topic we consider in this subsection is the current status of the field with regard to creating spiking LSTMs
or gated recurrent networks.

%Conventional recurrent neural networks (RNNs) such as fully connected RNNs~\cite{lecun2015deep,bengio1994learning}, long short term memory (LSTM)~\cite{hochreiter1997long}, and gated recurrent units (GRUs)~\cite{chung2014empirical} has shown remarkable performances in sequential data processing ~\cite{schmidhuber2015}. These neural architectures support the dynamic, temporal behavior of stimuli by interconnecting neurons to form a memory unit for sequential information. RNNs in SNNs can be categorized into two frameworks. First, by following the non-spiking RNN frameworks (e.g. LSTM) to develop recurrent SNNs. Second, using a large reservoir of spiking neurons (excitatory and inhibitory) to encode spatiotemporal information through interconnected spiking neurons. The first approach is more related to this review as we try to connect traditional deep learning to deep SNNs. 

There are only a few recurrent SNNs, in which conventional (non-spiking) RNNs are converted to spiking frameworks. 

Shrestha et al. (2017)~\cite{shrestha2017spike} 
implemented an energy-efficient spiking LSTM onto the IBM TrueNorth neurosynaptic system platform~\cite{akopyan2015truenorth}. 
To do this, they had to solve two problems.
The first was to build a spiking LSTM and the second was to implement it on a neuromorphic chip.
We focus on the former problem.
One of their design choices was to represent positive and negative values using two channels of spike trains.
This is bio-plausible and the DoG filters discussed in Section \ref{subsecDeepSpikingCNNs} commonly come in two forms for
exactly this purpose.
The inputs, outputs, and most of the internal LSTM variables used rate coding.
There was one exception to this where the value of the variable representing cell state
needed higher precision and was represented by a spike burst code.
The main thrust of the paper was overcoming the complications in mapping the LSTM to a neuromorphic
chip and accuracy results on standard benchmarks were not reported.

The phased LSTM \cite{Neil2016cPhased}, 
although not a spiking LSTM,
is well suited to process event-driven, asynchronously sampled data, which is a common task for SNNs.
This makes it useful to process inputs from (possibly several) biomorphic input sensors that sample inputs
at multiple time scales.
It is also potentially useful for processing outputs from SNNs.
The innovation of the phased LSTM is the addition of a time gate to the usual set of gates within a memory cell.
The time gate is synchronized to a rhythmic oscillation.
When `open', the time gate allows the usal updates to the hidden output and cell state vectors.
When `closed' it prevents the updates forcing the hidden and cell vectors to retain their values.
The units within these vectors can have separate with their own oscillation periods and phases.
This allows the phased LSTM to quantize its input at different time scales.
In a number of experiments that involved event-driven sampling at varied time scales,
the phased LSTM has trained more quickly than a regular LSTM while performing as accurately as
the regular LSTM\@.

%Another vein of research implements backpropagation through time (BPTT) in recent SNNs to improve recurrent SNN performance~\cite{huh2017gradient,bellec2018long}. 
%Bellec et. al. (2018)~\cite{bellec2018long} applied BPTT to a recurrent SNN consisting of a group of excitatory and inhibitory LIF neurons and a group of excitatory LIF neurons with decaying excitability through time to develop a spike-based LSTM. Their network showed high performance (matching a conventional LSTM) on sequential MNIST~\cite{le2015simple,costa2017cortical}. 

% Subsubsection
\subsubsection{Liquid State Machines and Reservoirs}
\label{LSMReservoir}

The neocortex, unique to mammals, has the ability to drastically scale its surface area from about 1 square cm in 
the mouse to about 2,500 square cm in the human
while keeping its thickness fairly constant ($\le$ 3 mm).
To support this expansion, one hypothesis is that mammals discovered a structural motif that may
be replicated and functionally adapted to new tasks.
Initially this was called a minicolumn consisting of about 300 excitatory and inhibitory recurrently connected
neurons that span the six layers of the three-millimeter-thick neocortex.
More recently the term canonical microcircuit has been used.
There has been great interest in modeling this hypothesized module.

In an effort to model the computations that might be taking place within the canonical neocortical microcircuit,
the liquid state machine (LSM) was introduced \cite{maass2002} 
which has since been partly absorbed by the field of reservoir computing \cite{lukovsevivcius2009reservoir,lukovsevivcius2012reservoir}.
In an LSM context, a neural reservoir is a sparsely connected recurrent SNN composed of excitatory and inhibitory neurons designed
to have enough structure to create a universal analog fading memory.
This module is constructed so that it can transform a set of possibly multi-modal spike trains into a spatiotemporal representation
whose instantaneous state can be recognized and readout by a layer of linear units.

A reservoir model has three parts:
\begin{enumerate}
\item  It needs one or more sensory-based, time-varying input streams of spikes or continuous inputs
that can be transduced into spikes.
\item It needs a recurrent SNN known as a reservoir or liquid whose synapses may (or may not) be able to learn.
      The neurons are given physical locations in space as a means to establish connection probabilities,
      which generally decrease exponentially with distance.
      Some ideas on why this might be useful are given in \cite{pyle2017spatiotemporal}.
      Connections tend to be sparse to avoid chaotic dynamics.
      The ratio of excitatory to inhibitory neurons is usually about 80\% to 20\% to reflect the ratio
      found in the neocortex.
\item It needs (usually) linear readout units which can be trained to recognize instantaneous patterns within the liquid.
      Part of the motivation for the linearity is to prove that the information in the reservoir's dynamically evolving
      state can be easily read out.
\end{enumerate}

%A reservoir of spiking neurons (excitatory and inhibitory neurons) resembles a recurrent SNN that captures the spatiotemporal behavior of input spike trains. 
%Fortunately, a rich track of investigations has been fulfilled to perform reliable dynamic computations in reservoir computing frameworks~\cite{lukovsevivcius2009reservoir,lukovsevivcius2012reservoir}.
The neuromorphic cube (NewCube) \cite{kasabov2014neucube,paulun2018retinotopic,schliebs2011}
is a broad and futuristic model proposed as a unifying computational architecture for
modeling multi-modality spatiotemporal data, most especially data related to the brain,
such as EEG analysis \cite{doborjeh2018modelling}.
The core of the NeuCube architecture is a 3D reservoir of spiking neurons trained by an STDP-like
mechanism.
The reservoir's structure is intended to directly reflect the inter-area connectivity of the human
neocortex as revealed by structural connectivity based on anatomy such as diffusion tensor imaging (DTI)
and functional  connectivity based on measures like functional magnetic resonance imaging (fMRI)\@.
This is contrast to the use of a reservoir as a model of a neocortical microcircuit.
The claim of NeuCube is that the connectivity of the brain at the macro scale obeys the properties of a reservoir.

The previous subsection discussed the status of attempts to create spiking versions of LSTMs.
Rather than pursuing a direct approach to structurally translating an LSTM to a spiking version,
the work of \cite{bellec2018long} 
took a reservoir inspired approach.
Their LSNN architecture (long short-term memory SNNs) consisted of four modules labeled:
$X$, $R$, $A$, and $Y$.
$X$ provided multiple streams of input spikes, $R$ was the reservoir consisting of excitatory and inhibitory
neurons, $A$ was a module of excitatory neurons (connected to $X$, $R$, and $Y$) with adaptive thresholds whose
purpose was in part to maintain a tight excitatory-inhibitory balance in $R$, and module $Y$ 
consisted of the
readout neurons.
The LSNN was trained using BPTT (Section \ref{subsubGatedSNNs}) 
with pseudo derivatives using the membrane potential (as explained in Section \ref{subsubSupervisedLearning}).
The network achieved comparable accuracy performance to LSTMs on the sequential MNIST benchmark and also on the TIMIT Acoustic-Phonetic Continuous Speech Corpus.
Sequencial MNIST is a sequence learning benchmark for assessing recurrent network performance first described in \cite{Lamb2016a}.
The task is to recognize MNIST digits but now the input is the set of $784=28^2$ input pixels
delivered sequentially over consecutive time steps.
Although the LSNN architecture does not have a direct mapping to the architecture of an LSTM,
it has learning abilities that are apparently unique to LSTMs.
It has been shown that LSTMs ``can learn nonlinear functions from a teacher without modifying their weights,
and using their short-term memory instead.'' In \cite{Hochreiter2001a}, it was shown that LSTMs had this property.

Another recent study, \cite{costa2017cortical}, 
has attempted to adapt the architecture of a conventional LSTM to
be a plausible model of a cortical microcircuit.
There were two innovations in this model.
First, to facilitate mapping to a microcircuit, the multiplicative gating operations within a standard LSTM were
replaced with subtractive operations that could be implemented by lateral inhibitory circuits in the neocortex.
This lead to the name subLSTM (subtractive LSTM).
Second, to facilitate learning and study using readily available deep learning frameworks, the neural coding scheme
was assumed to use rate-coded LIF neurons.
This allowed them to model spiking neurons using a continuous approximation that was compatible with deep learning environments.
Their bio-plausible subLSTM achieved comparable performance to a standard LSTM on the sequential MNIST task.
Similar results were obtained in a language processing benchmark.
The subLSTMS did not perform better than standard LSTMs but they have opened an avenue for interdisciplinary
dialog between questions of brain function and theoretical insights from deep learning.

%Using the neuromorphic cube (NeuCube)~\cite{kasabov2014neucube,paulun2018retinotopic,schliebs2011} for simulating brain-like computations, 
%and distance-based connection probability in the reservoir~\cite{pyle2017spatiotemporal} are recent examples of this type of the recurrent SNN architecture. 
%Most recently, a recurrent SNN was developed using the NeuCube framework to analyze EEG readouts of pre-perceptual perception in the brain~\cite{doborjeh2018modelling}. 

These studies show a bright future of recurrent SNNs which take advantages of the conventional RNNs and the bio-inspired recurrent neural framework of reservoir computing.  

\subsection{Performance Comparisons of Contemporary Models}

\begin{table*}
\footnotesize
\setlength\tabcolsep{2pt}
\centering
\caption{Summary of recent deep learning models developed in SNN platforms and their accuracy on MNIST~\cite{lecun1998gradient,lecun1998mnist}, N-MNIST~\cite{orchard2015converting}, CIFAR-10, and CIFAR-100~\cite{krizhevsky2009learning}. MNIST: handwritten digits. N-MNIST: neuromorphic-MNIST representing a spiking version of MNIST. CIFAR: tiny colored images.}
\label{tab:summary}
\begin{tabular}{|l|l|l|l|l|}
\hline
\multicolumn{1}{|c|}{Model}          & \multicolumn{1}{c|}{Architecture} & \multicolumn{1}{c|}{Learning method}                         & \multicolumn{1}{c|}{Dataset} & \multicolumn{1}{c|}{Acc} \\ \hline
\multicolumn{5}{|c|}{\textbf{Feedforward, fully connected, multi-layer SNNs}}                                                                                                                                                            \\ \hline
O'Connor (2016)~\cite{o2016deep}       & Deep SNN                          & Stochastic gradient descent                           & MNIST                        & 96.40                                   \\ \hline
O'Connor (2016)~\cite{o2016deep}       & Deep SNN                          & Fractional stochastic gradient descent                & MNIST                        & 97.93                                   \\ \hline
Lee (2016)~\cite{lee2016training}            & Deep SNN                          & Backpropagation                                       & MNIST                        & 98.88                                   \\ \hline
Lee (2016)~\cite{lee2016training}            & Deep SNN                          & Backpropagation                                       & N-MNIST                      & 98.74                                   \\ \hline
Neftci (2017)~\cite{neftci2017event}            & Deep SNN                          & Event-driven random backpropagation                                       & MNIST                      & 97.98                                   \\ \hline
Liu (2017)~\cite{liu2017mt}         & SNN                               & Temporal backpropagation (3-layer)                                                  & MNIST                        & 99.10                                   \\ \hline
Eliasmith (2012)~\cite{eliasmith2012large}            & SNN                               & Spaun brain model                                     & MNIST                        & 94.00                                   \\ \hline
Diehl (2015)~\cite{diehl2015unsupervised}         & SNN                               & STDP (2-layer)                                                  & MNIST                        & 95.00                                   \\ \hline
Tavanaei (2017)~\cite{tavanaei2017bp}         & SNN                               & STDP-based backpropagation (3-layer)                                                  & MNIST                        & 97.20                                   \\ \hline
Mostafa (2017)~\cite{mostafa2017supervised}         & SNN                               & Temporal backpropagation (3-layer)                                                  & MNIST                        & 97.14                                   \\ \hline
Querlioz (2013)~\cite{querlioz2013immunity}      & SNN                               & STDP, Hardware implementation                         & MNIST                        & 93.50                                   \\ \hline
Brader (2007)\cite{brader2007learning}        & SNN                               & Spike-driven synaptic plasticity                      & MNIST                        & 96.50                                   \\ \hline
Diehl (2015)~\cite{diehl2015}          & Deep SNN                          & Offline learning, Conversion                          & MNIST                        & 98.60                                   \\ \hline
Neil (2016)~\cite{neil2016learning}           & Deep SNN                          & Offline learning, Conversion                          & MNIST                        & 98.00                                   \\ \hline
Hunsberger (2015)~\cite{hunsberger2015spiking,hunsberger2016training} & Deep SNN                          & Offline learning, Conversion                          & MNIST                        & 98.37                                   \\ \hline
Esser (2015)~\cite{esser2015backpropagation}         & Deep SNN                          & Offline learning, Conversion                          & MNIST                        & 99.42                                   \\ \hline
\multicolumn{5}{|c|}{\textbf{Spiking CNNs}}                                                                                                                                                               \\ \hline
Lee (2016)~\cite{lee2016training}            & Spiking CNN                       & Backpropagation                                       & MNIST                        & 99.31                                   \\ \hline
Lee (2016)~\cite{lee2016training}            & Spiking CNN                       & Backpropagation                                       & N-MNIST                      & 98.30                                   \\ \hline
Panda (2016)~\cite{panda2016unsupervised}      & Spiking CNN                       & Convolutional autoencoder                             & MNIST                        & 99.05                                   \\ \hline
Panda (2016)~\cite{panda2016unsupervised}     & Spiking CNN                       & Convolutional autoencoder                             & CIFAR-10                     & 75.42                                   \\ \hline
Tavanaei (2017)~\cite{tavanaei2016bio,tavanaei2017multi}   & Spiking CNN                       & Layer wise sparse coding and STDP                     & MNIST                        & 98.36                                   \\ \hline
Tavanaei (2018)~\cite{tavanaei2018training}   & Spiking CNN                       & Layer-wise and end-to-end STDP rules                     & MNIST                        & 98.60                                   \\ \hline
Kheradpisheh (2016)~\cite{kheradpisheh2016stdp}  & Spiking CNN                       & Layer wise STDP                                       & MNIST                        & 98.40                                   \\ \hline
Zhao (2015)~\cite{zhao2015feedforward}          & Spiking CNN                       & Tempotron                                             & MNIST                        & 91.29                                   \\ \hline
Cao (2015)~\cite{cao2015}         & Spiking CNN                       & Offline learning, Conversion                          & CIFAR-10                     & 77.43                                   \\ \hline
Neil (2016)~\cite{neil2016effective}          & Spiking CNN                       & Offline learning, Conversion                          & N-MNIST                      & 95.72                                   \\ \hline
Diehl (2015)~\cite{diehl2015}          & Spiking CNN                       & Offline learning, Conversion                          & MNIST                        & 99.10                                   \\ \hline
Rueckauer (2017)~\cite{rueckauer2017conversion}      & Spiking CNN                       & Offline learning, Conversion                          & MNIST                        & 99.44                                   \\ \hline
Rueckauer (2017)~\cite{rueckauer2017conversion}      & Spiking CNN                       & Offline learning, Conversion                          & CIFAR-10                     & 90.85                                   \\ \hline
Hunsberger (2015)~\cite{hunsberger2015spiking} & Spiking CNN                       & Offline learning, Conversion                          & CIFAR-10                     & 82.95                                   \\ \hline
Garbin (2014)~\cite{garbin2014variability}        & Spiking CNN                       & Offline learning, Hardware & MNIST                        & 94.00                                   \\ \hline
Esser (2016)~\cite{esser2016convolutional}         & Spiking CNN                       & Offline learning, Hardware & CIFAR-10                     & 87.50                                   \\ \hline
Esser (2016)~\cite{esser2016convolutional}         & Spiking CNN                       & Offline learning, Hardware & CIFAR-100                    & 63.05                                   \\ \hline
\multicolumn{5}{|c|}{\textbf{Spiking RBMs and DBNs}}                                                                                                                                                      \\ \hline
Neftci (2014)~\cite{neftci2014}    & Spiking RBM                       & Contrastive divergence in LIF neurons                 & MNIST                        & 91.90                                   \\ \hline
O'Connor (2013)~\cite{o2013real}       & Spiking DBN                       & Offline learning, Conversion                          & MNIST                        & 94.09                                   \\ \hline
Stromatias (2015)~\cite{stromatias2015robustness}     & Spiking DBN                       & Offline learning, Conversion                          & MNIST                        & 94.94                                   \\ \hline
Stromatias (2015)~\cite{stromatias2015scalable}     & Spiking DBN                       & Offline learning, Hardware & MNIST                        & 95.00                                   \\ \hline
Merolla (2011)~\cite{merolla2011digital}       & Spiking RBM                       & Offline learning, Hardware & MNIST                        & 94.00                                   \\ \hline
Neil (2014)~\cite{neil2014minitaur}          & Spiking DBN                       & Offline learning, Hardware & MNIST                        & 92.00                                   \\ \hline
\end{tabular}
\end{table*} 

Table~\ref{tab:summary} shows the previous models for developing deep SNNs and their architectures along with their accuracy rates on different datasets. This table shows two tracks of spiking models: 1) using online learning and 2) using offline learning (deployment). The latter method has reported higher performance but it avoids training the multi-layer SNNs by converting the offline trained neural networks to the relevant spiking platform. On the other hand, online learning offers multi-layer learning in SNNs but reports lower accuracy rates. Additionally, as expected, the spiking CNNs have achieved higher accuracy rates than the spiking DBNs and the fully connected SNNs on image classification. This comparison provides insight into different SNN architectures and learning mechanisms to choose the right tool for the right purpose in future investigations.

\section{Summary}
Deep learning approaches have recently shown breakthrough performance in many areas of pattern recognition recently. In spite of their effectiveness in hierarchical feature extraction and classification, these types of neural networks are computationally expensive and difficult to implement on hardware for portable devices. In another line of research of neural network architectures, SNNs have been described as power-efficient models because of their sparse, spike-based communication framework. 
%SNNs are brain-inspired and biologically plausible such that they seek to mimic the accurate and efficient functionality of brain.
Recent studies try to take advantages of both frameworks (deep learning and SNN) to develop a multi-layer SNN architecture to achieve high performance of recently proved deep networks while implementing bio-inspired, power-efficient platforms. Additionally, the literature has shown that the brain detects stimuli patterns through multi-layer SNNs communicating by spike trains via adaptive synapses. The biologically realistic spiking neurons communicate using spike trains which do not have obvious derivatives. This makes SNNs unable to directly use derivative-based optimization for training. This paper reviewed novel learning approaches for different layers of SNNs to address some of the open questions in this field. As biological neurons use sparse, stochastic, spike-based communication, a spiking network can be an appropriate starting point for modeling the brain's functionality.

SNNs with specific neural architectures demand new neuron models and learning techniques. Spiking neurons communicate through discrete spike trains via synapses adapting locally to distinguish the pattern of stimuli. The quest to meet these requirements can be accomplished by bio-inspired neural simulations for integrating the stimuli and releasing discriminative spike patterns according to the adaptive filters associated with synaptic weight sets. An important challenge in developing SNNs is to develop appropriate learning rules to detect spatio-temporally local patterns of spike trains. 
In this paper we reviewed state-of-the-art deep SNNs developed to reach the performance of conventional deep learning methods while providing a bio-inspired, power-efficient platform. Three popular deep learning methods as deep fully connected SNNs, spiking CNNs, and spiking DBNs were reviewed. The performances reported by recent approaches determine that the spike-based deep learning methods perform as well as traditional DNNs. Furthermore, SNNs are based on the human brain functionality and are able to perform much better than traditional ones in the future, as human brain does. This paper reviewed methods, network architectures, experiments, and results of recently proposed deep spiking networks to be useful for next studies and experiments.       

%\section*{References}
\bibliographystyle{IEEEtran} 
%\bibliographystyle{unsrt}
\bibliography{amir}

\begin{thebibliography}{100}
\providecommand{\url}[1]{#1}
\csname url@samestyle\endcsname
\providecommand{\newblock}{\relax}
\providecommand{\bibinfo}[2]{#2}
\providecommand{\BIBentrySTDinterwordspacing}{\spaceskip=0pt\relax}
\providecommand{\BIBentryALTinterwordstretchfactor}{4}
\providecommand{\BIBentryALTinterwordspacing}{\spaceskip=\fontdimen2\font plus
\BIBentryALTinterwordstretchfactor\fontdimen3\font minus
  \fontdimen4\font\relax}
\providecommand{\BIBforeignlanguage}[2]{{%
\expandafter\ifx\csname l@#1\endcsname\relax
\typeout{** WARNING: IEEEtran.bst: No hyphenation pattern has been}%
\typeout{** loaded for the language `#1'. Using the pattern for}%
\typeout{** the default language instead.}%
\else
\language=\csname l@#1\endcsname
\fi
#2}}
\providecommand{\BIBdecl}{\relax}
\BIBdecl

\bibitem{krizhevsky2012}
A.~Krizhevsky, I.~Sutskever, and G.~E. Hinton, ``Imagenet classification with
  deep convolutional neural networks,'' in \emph{Advances in neural information
  processing systems}, 2012, pp. 1097--1105.

\bibitem{ILSVRC15}
O.~Russakovsky, J.~Deng, H.~Su, J.~Krause, S.~Satheesh, S.~Ma, Z.~Huang,
  A.~Karpathy, A.~Khosla, M.~Bernstein, A.~C. Berg, and L.~Fei-Fei, ``{ImageNet
  Large Scale Visual Recognition Challenge},'' \emph{International Journal of
  Computer Vision (IJCV)}, vol. 115, no.~3, pp. 211--252, 2015.

\bibitem{lecun2015deep}
Y.~LeCun, Y.~Bengio, and G.~Hinton, ``Deep learning,'' \emph{Nature}, vol. 521,
  no. 7553, pp. 436--444, 2015.

\bibitem{schmidhuber2015}
J.~Schmidhuber, ``Deep learning in neural networks: An overview,'' \emph{Neural
  Networks}, vol.~61, pp. 85--117, 2015.

\bibitem{szegedy2016rethinking}
C.~Szegedy, V.~Vanhoucke, S.~Ioffe, J.~Shlens, and Z.~Wojna, ``Rethinking the
  inception architecture for computer vision,'' in \emph{Proceedings of the
  IEEE Conference on Computer Vision and Pattern Recognition}, 2016, pp.
  2818--2826.

\bibitem{he2016deep}
K.~He, X.~Zhang, S.~Ren, and J.~Sun, ``Deep residual learning for image
  recognition,'' in \emph{Proceedings of the IEEE conference on computer vision
  and pattern recognition}, 2016, pp. 770--778.

\bibitem{long2015fully}
J.~Long, E.~Shelhamer, and T.~Darrell, ``Fully convolutional networks for
  semantic segmentation,'' in \emph{Proceedings of the IEEE Conference on
  Computer Vision and Pattern Recognition}, 2015, pp. 3431--3440.

\bibitem{girshick2014rich}
R.~Girshick, J.~Donahue, T.~Darrell, and J.~Malik, ``Rich feature hierarchies
  for accurate object detection and semantic segmentation,'' in
  \emph{Proceedings of the IEEE conference on computer vision and pattern
  recognition}, 2014, pp. 580--587.

\bibitem{hinton2012deep}
G.~Hinton, L.~Deng, D.~Yu, G.~E. Dahl, A.-r. Mohamed, N.~Jaitly, A.~Senior,
  V.~Vanhoucke, P.~Nguyen, T.~N. Sainath \emph{et~al.}, ``Deep neural networks
  for acoustic modeling in speech recognition: The shared views of four
  research groups,'' \emph{IEEE Signal Processing Magazine}, vol.~29, no.~6,
  pp. 82--97, 2012.

\bibitem{mamoshina2016applications}
P.~Mamoshina, A.~Vieira, E.~Putin, and A.~Zhavoronkov, ``Applications of deep
  learning in biomedicine,'' \emph{Molecular Pharmaceutics}, vol.~13, no.~5,
  pp. 1445--1454, 2016.

\bibitem{min2017deep}
S.~Min, B.~Lee, and S.~Yoon, ``Deep learning in bioinformatics,''
  \emph{Briefings in bioinformatics}, vol.~18, no.~5, pp. 851--869, 2017.

\bibitem{venna2017novel}
S.~R. Venna, A.~Tavanaei, R.~N. Gottumukkala, V.~V. Raghavan, A.~Maida, and
  S.~Nichols, ``A novel data-driven model for real-time influenza
  forecasting,'' \emph{bioRxiv}, p. 185512, 2017.

\bibitem{hassabis2017neuroscience}
D.~Hassabis, D.~Kumaran, C.~Summerfield, and M.~Botvinick,
  ``Neuroscience-inspired artificial intelligence,'' \emph{Neuron}, vol.~95,
  no.~2, pp. 245--258, 2017.

\bibitem{vanrullen2017perception}
R.~VanRullen, ``Perception science in the age of deep neural networks,''
  \emph{Frontiers in Psychology}, vol.~8, p. 142, 2017.

\bibitem{gerstner2014neuronal}
W.~Gerstner, W.~M. Kistler, R.~Naud, and L.~Paninski, \emph{Neuronal dynamics:
  From single neurons to networks and models of cognition}.\hskip 1em plus
  0.5em minus 0.4em\relax Cambridge University Press, 2014.

\bibitem{hodgkin1952quantitative}
A.~L. Hodgkin and A.~F. Huxley, ``A quantitative description of membrane
  current and its application to conduction and excitation in nerve,''
  \emph{The Journal of physiology}, vol. 117, no.~4, pp. 500--544, 1952.

\bibitem{Sejnowski1995a}
Z.~F. Mainen and T.~J. Sejnowski, ``Reliability of spike timing in neocortical
  neurons,'' \emph{Science}, vol. 268, pp. 1503--1506, 1995.

\bibitem{Bair1996a}
W.~Bair and C.~Koch, ``Temporal precision of spike trains in extrastriate
  cortex of the behaving macaque monkey,'' \emph{Neural Computation}, vol.~8,
  no.~6, pp. 1185--1202, 1996.

\bibitem{Herikstad2011a}
R.~Herikstad, J.~Baker, J.~Lachaux, C.~Gray, and S.~Yen, ``Natural movies evoke
  spike trains with low spike time variability in cat primary visual cortex,''
  \emph{Journal of Neuroscience}, vol.~31, no.~44, pp. 15\,844--15\,860, 2011.

\bibitem{gollisch2008rapid}
T.~Gollisch and M.~Meister, ``Rapid neural coding in the retina with relative
  spike latencies,'' \emph{Science}, vol. 319, no. 5866, pp. 1108--1111, 2008.

\bibitem{sinha2017cellular}
R.~Sinha, M.~Hoon, J.~Baudin, H.~Okawa, R.~O. Wong, and F.~Rieke, ``Cellular
  and circuit mechanisms shaping the perceptual properties of the primate
  fovea,'' \emph{Cell}, vol. 168, no.~3, pp. 413--426, 2017.

\bibitem{victor2005spike}
J.~D. Victor, ``Spike train metrics,'' \emph{Current opinion in neurobiology},
  vol.~15, no.~5, pp. 585--592, 2005.

\bibitem{butts2007temporal}
D.~A. Butts, C.~Weng, J.~Jin, C.-I. Yeh, N.~A. Lesica, J.-M. Alonso, and G.~B.
  Stanley, ``Temporal precision in the neural code and the timescales of
  natural vision,'' \emph{Nature}, vol. 449, no. 7158, pp. 92--95, 2007.

\bibitem{reinagel2000temporal}
P.~Reinagel and R.~C. Reid, ``Temporal coding of visual information in the
  thalamus,'' \emph{Journal of Neuroscience}, vol.~20, no.~14, pp. 5392--5400,
  2000.

\bibitem{srivastava2017motor}
K.~H. Srivastava, C.~M. Holmes, M.~Vellema, A.~R. Pack, C.~P. Elemans,
  I.~Nemenman, and S.~J. Sober, ``Motor control by precisely timed spike
  patterns,'' \emph{Proceedings of the National Academy of Sciences}, p.
  201611734, 2017.

\bibitem{tang2014millisecond}
C.~Tang, D.~Chehayeb, K.~Srivastava, I.~Nemenman, and S.~J. Sober,
  ``Millisecond-scale motor encoding in a cortical vocal area,'' \emph{PLoS
  biology}, vol.~12, no.~12, p. e1002018, 2014.

\bibitem{wysoski2010}
S.~G. Wysoski, L.~Benuskova, and N.~Kasabov, ``Evolving spiking neural networks
  for audiovisual information processing,'' \emph{Neural Networks}, vol.~23,
  no.~7, pp. 819--835, 2010.

\bibitem{gupta2007}
A.~Gupta and L.~N. Long, ``Character recognition using spiking neural
  networks,'' in \emph{Neural Networks, 2007. IJCNN 2007. International Joint
  Conference on}.\hskip 1em plus 0.5em minus 0.4em\relax IEEE, 2007, pp.
  53--58.

\bibitem{meftah2010}
B.~Meftah, O.~Lezoray, and A.~Benyettou, ``Segmentation and edge detection
  based on spiking neural network model,'' \emph{Neural Processing Letters},
  vol.~32, no.~2, pp. 131--146, 2010.

\bibitem{escobar2009}
M.-J. Escobar, G.~S. Masson, T.~Vieville, and P.~Kornprobst, ``Action
  recognition using a bio-inspired feedforward spiking network,''
  \emph{International Journal of Computer Vision}, vol.~82, no.~3, pp.
  284--301, 2009.

\bibitem{liaw1998}
J.-S. Liaw and T.~W. Berger, ``Robust speech recognition with dynamic
  synapses,'' in \emph{Neural Networks Proceedings, 1998. IEEE World Congress
  on Computational Intelligence. The 1998 IEEE International Joint Conference
  on}, vol.~3.\hskip 1em plus 0.5em minus 0.4em\relax IEEE, 1998, pp.
  2175--2179.

\bibitem{kroger2009}
B.~J. Kr{\"o}ger, J.~Kannampuzha, and C.~Neuschaefer-Rube, ``Towards a
  neurocomputational model of speech production and perception,'' \emph{Speech
  Communication}, vol.~51, no.~9, pp. 793--809, 2009.

\bibitem{panchev2004}
C.~Panchev and S.~Wermter, ``Spike-timing-dependent synaptic plasticity: from
  single spikes to spike trains,'' \emph{Neurocomputing}, vol.~58, pp.
  365--371, 2004.

\bibitem{namarvar2001}
A.~Tavanaei and A.~Maida, ``Bio-inspired multi-layer spiking neural network
  extracts discriminative features from speech signals,'' in
  \emph{International Conference on Neural Information Processing}.\hskip 1em
  plus 0.5em minus 0.4em\relax Springer, 2017, pp. 899--908.

\bibitem{wade2010}
J.~J. Wade, L.~J. McDaid, J.~A. Santos, and H.~M. Sayers, ``{SWAT}: a spiking
  neural network training algorithm for classification problems,'' \emph{Neural
  Networks, IEEE Transactions on}, vol.~21, no.~11, pp. 1817--1830, 2010.

\bibitem{nager2002}
C.~N{\"a}ger, J.~Storck, and G.~Deco, ``Speech recognition with spiking neurons
  and dynamic synapses: a model motivated by the human auditory pathway,''
  \emph{Neurocomputing}, vol.~44, pp. 937--942, 2002.

\bibitem{loiselle2005}
S.~Loiselle, J.~Rouat, D.~Pressnitzer, and S.~Thorpe, ``Exploration of rank
  order coding with spiking neural networks for speech recognition,'' in
  \emph{Neural Networks, 2005. IJCNN'05. Proceedings. 2005 IEEE International
  Joint Conference on}, vol.~4.\hskip 1em plus 0.5em minus 0.4em\relax IEEE,
  2005, pp. 2076--2080.

\bibitem{ghosh2007}
S.~Ghosh-Dastidar and H.~Adeli, ``Improved spiking neural networks for {EEG}
  classification and epilepsy and seizure detection,'' \emph{Integrated
  Computer-Aided Engineering}, vol.~14, no.~3, pp. 187--212, 2007.

\bibitem{kasabov2014b}
N.~Kasabov, V.~Feigin, Z.-G. Hou, Y.~Chen, L.~Liang, R.~Krishnamurthi,
  M.~Othman, and P.~Parmar, ``Evolving spiking neural networks for personalised
  modelling, classification and prediction of spatio-temporal patterns with a
  case study on stroke,'' \emph{Neurocomputing}, vol. 134, pp. 269--279, 2014.

\bibitem{felleman1991distributed}
D.~J. Felleman and D.~C. Van~Essen, ``Distributed hierarchical processing in
  the primate cerebral cortex.'' \emph{Cerebral Cortex (New York, NY: 1991)},
  vol.~1, no.~1, pp. 1--47, 1991.

\bibitem{serre2014hierarchical}
T.~Serre, ``Hierarchical models of the visual system,'' in \emph{Encyclopedia
  of computational neuroscience}.\hskip 1em plus 0.5em minus 0.4em\relax
  Springer, 2014, pp. 1--12.

\bibitem{freiwald2010functional}
W.~A. Freiwald and D.~Y. Tsao, ``Functional compartmentalization and viewpoint
  generalization within the macaque face-processing system,'' \emph{Science},
  vol. 330, no. 6005, pp. 845--851, 2010.

\bibitem{Stone2018a}
J.~V. Stone, \emph{Principles of Neural Information Theory: Computational
  Neuroscience and Metabolic Efficiency}.\hskip 1em plus 0.5em minus
  0.4em\relax Sebtel Press, 2018.

\bibitem{merolla2014million}
P.~A. Merolla, J.~V. Arthur, R.~Alvarez-Icaza, A.~S. Cassidy, J.~Sawada,
  F.~Akopyan, B.~L. Jackson, N.~Imam, C.~Guo, Y.~Nakamura \emph{et~al.}, ``A
  million spiking-neuron integrated circuit with a scalable communication
  network and interface,'' \emph{Science}, vol. 345, no. 6197, pp. 668--673,
  2014.

\bibitem{seo201145nm}
J.-s. Seo, B.~Brezzo, Y.~Liu, B.~D. Parker, S.~K. Esser, R.~K. Montoye,
  B.~Rajendran, J.~A. Tierno, L.~Chang, D.~S. Modha \emph{et~al.}, ``A 45nm
  {CMOS} neuromorphic chip with a scalable architecture for learning in
  networks of spiking neurons,'' in \emph{Custom Integrated Circuits Conference
  (CICC), 2011 IEEE}.\hskip 1em plus 0.5em minus 0.4em\relax IEEE, 2011, pp.
  1--4.

\bibitem{carrillo2012advancing}
S.~Carrillo, J.~Harkin, L.~McDaid, S.~Pande, S.~Cawley, B.~McGinley, and
  F.~Morgan, ``Advancing interconnect density for spiking neural network
  hardware implementations using traffic-aware adaptive network-on-chip
  routers,'' \emph{Neural networks}, vol.~33, pp. 42--57, 2012.

\bibitem{carrillo2013scalable}
S.~Carrillo, J.~Harkin, L.~J. McDaid, F.~Morgan, S.~Pande, S.~Cawley, and
  B.~McGinley, ``Scalable hierarchical network-on-chip architecture for spiking
  neural network hardware implementations,'' \emph{IEEE Transactions on
  Parallel and Distributed Systems}, vol.~24, no.~12, pp. 2451--2461, 2013.

\bibitem{bengio2009learning}
Y.~Bengio, ``Learning deep architectures for {AI},'' \emph{Foundations and
  Trends in Machine Learning}, vol.~2, no.~1, pp. 1--127, 2009.

\bibitem{maass2015spike}
W.~Maass, ``To spike or not to spike: that is the question,'' \emph{Proceedings
  of the IEEE}, vol. 103, no.~12, pp. 2219--2224, 2015.

\bibitem{maass1997}
------, ``Networks of spiking neurons: the third generation of neural network
  models,'' \emph{Neural networks}, vol.~10, no.~9, pp. 1659--1671, 1997.

\bibitem{rozenberg2011handbook}
G.~Rozenberg, T.~Bck, and J.~N. Kok, \emph{Handbook of natural
  computing}.\hskip 1em plus 0.5em minus 0.4em\relax Springer Publishing
  Company, Incorporated, 2011.

\bibitem{seung2003learning}
H.~S. Seung, ``Learning in spiking neural networks by reinforcement of
  stochastic synaptic transmission,'' \emph{Neuron}, vol.~40, no.~6, pp.
  1063--1073, 2003.

\bibitem{liu2005repeated}
Q.-s. Liu, L.~Pu, and M.-m. Poo, ``Repeated cocaine exposure in vivo
  facilitates {LTP} induction in midbrain dopamine neurons,'' \emph{Nature},
  vol. 437, no. 7061, p. 1027, 2005.

\bibitem{song2013asynchronous}
T.~Song, L.~Pan, and G.~P{\u{a}}un, ``Asynchronous spiking neural p systems
  with local synchronization,'' \emph{Information Sciences}, vol. 219, pp.
  197--207, 2013.

\bibitem{chavez1990decrease}
L.~Chavez-Noriega, J.~Halliwell, and T.~Bliss, ``A decrease in firing threshold
  observed after induction of the epsp-spike (es) component of long-term
  potentiation in rat hippocampal slices,'' \emph{Experimental brain research},
  vol.~79, no.~3, pp. 633--641, 1990.

\bibitem{huh2017gradient}
D.~Huh and T.~J. Sejnowski, ``Gradient descent for spiking neural networks,''
  \emph{arXiv preprint arXiv:1706.04698}, pp. 1--10, 2017.

\bibitem{lee2016training}
J.~H. Lee, T.~Delbruck, and M.~Pfeiffer, ``Training deep spiking neural
  networks using backpropagation,'' \emph{Frontiers in Neuroscience}, vol.~10,
  p. 508, 2016.

\bibitem{Maass1996a}
W.~Maass, ``Lower bounds for the computational power of networks of spiking
  neurons.'' \emph{Neural Computation}, vol.~8, no.~1, pp. 1--40, 1996.

\bibitem{masquelier2007}
T.~Masquelier and S.~J. Thorpe, ``Unsupervised learning of visual features
  through spike timing dependent plasticity,'' \emph{PLoS Comput Biol}, vol.~3,
  no.~2, p. e31, 2007.

\bibitem{tavanaei2016c}
A.~Tavanaei, T.~Masquelier, and A.~S. Maida, ``Acquisition of visual features
  through probabilistic spike timing dependent plasticity,'' in \emph{Neural
  Networks (IJCNN), The 2016 International Joint Conference on}.\hskip 1em plus
  0.5em minus 0.4em\relax IEEE, 2016, pp. 1--8.

\bibitem{beyeler2013}
M.~Beyeler, N.~D. Dutt, and J.~L. Krichmar, ``Categorization and
  decision-making in a neurobiologically plausible spiking network using a
  {STDP}-like learning rule,'' \emph{Neural Networks}, vol.~48, pp. 109--124,
  2013.

\bibitem{ghosh2009b}
S.~Ghosh-Dastidar and H.~Adeli, ``Spiking neural networks,''
  \emph{International Journal of Neural Systems}, vol.~19, no.~04, pp.
  295--308, 2009.

\bibitem{kasabov2013}
N.~Kasabov, K.~Dhoble, N.~Nuntalid, and G.~Indiveri, ``Dynamic evolving spiking
  neural networks for on-line spatio-and spectro-temporal pattern
  recognition,'' \emph{Neural Networks}, vol.~41, pp. 188--201, 2013.

\bibitem{gerstner2002spiking}
W.~Gerstner and W.~M. Kistler, \emph{Spiking neuron models: Single neurons,
  populations, plasticity}.\hskip 1em plus 0.5em minus 0.4em\relax Cambridge
  University Press, 2002.

\bibitem{rieke1999spikes}
F.~Rieke, \emph{Spikes: exploring the neural code}.\hskip 1em plus 0.5em minus
  0.4em\relax MIT press, 1999.

\bibitem{bohte2004evidence}
S.~M. Bohte, ``The evidence for neural information processing with precise
  spike-times: A survey,'' \emph{Natural Computing}, vol.~3, no.~2, pp.
  195--206, 2004.

\bibitem{hopfield1995pattern}
J.~J. Hopfield \emph{et~al.}, ``Pattern recognition computation using action
  potential timing for stimulus representation,'' \emph{Nature}, vol. 376, no.
  6535, pp. 33--36, 1995.

\bibitem{bohte2002unsupervised}
S.~M. Bohte, H.~La~Poutr{\'e}, and J.~N. Kok, ``Unsupervised clustering with
  spiking neurons by sparse temporal coding and multilayer {RBF} networks,''
  \emph{IEEE Transactions on neural networks}, vol.~13, no.~2, pp. 426--435,
  2002.

\bibitem{kistler1997reduction}
W.~M. Kistler, W.~Gerstner, and J.~L. van Hemmen, ``Reduction of the
  {Hodgkin-Huxley} equations to a single-variable threshold model,''
  \emph{Neural Computation}, vol.~9, no.~5, pp. 1015--1045, 1997.

\bibitem{jolivet2003spike}
R.~Jolivet, J.~Timothy, and W.~Gerstner, ``The spike response model: a
  framework to predict neuronal spike trains,'' in \emph{Artificial Neural
  Networks and Neural Information Processing---ICANN/ICONIP 2003}.\hskip 1em
  plus 0.5em minus 0.4em\relax Springer, 2003, pp. 846--853.

\bibitem{izhikevich2003simple}
E.~M. Izhikevich \emph{et~al.}, ``Simple model of spiking neurons,'' \emph{IEEE
  Transactions on neural networks}, vol.~14, no.~6, pp. 1569--1572, 2003.

\bibitem{delorme1999spikenet}
A.~Delorme, J.~Gautrais, R.~Van~Rullen, and S.~Thorpe, ``Spikenet: A simulator
  for modeling large networks of integrate and fire neurons,''
  \emph{Neurocomputing}, vol.~26, pp. 989--996, 1999.

\bibitem{kandel2000principles}
E.~R. Kandel, J.~H. Schwartz, T.~M. Jessell, S.~A. Siegelbaum, A.~J. Hudspeth
  \emph{et~al.}, \emph{Principles of neural science}.\hskip 1em plus 0.5em
  minus 0.4em\relax McGraw-hill New York, 2000, vol.~4.

\bibitem{caporale2008spike}
N.~Caporale and Y.~Dan, ``Spike timing-dependent plasticity: a {Hebbian}
  learning rule,'' \emph{Annu. Rev. Neurosci.}, vol.~31, pp. 25--46, 2008.

\bibitem{markram2011history}
H.~Markram, W.~Gerstner, and P.~J. Sj{\"o}str{\"o}m, ``A history of
  spike-timing-dependent plasticity,'' \emph{Spike-timing dependent
  plasticity}, p.~11, 2011.

\bibitem{dan2006spike}
Y.~Dan and M.-M. Poo, ``Spike timing-dependent plasticity: from synapse to
  perception,'' \emph{Physiological reviews}, vol.~86, no.~3, pp. 1033--1048,
  2006.

\bibitem{song2000competitive}
S.~Song, K.~D. Miller, and L.~F. Abbott, ``Competitive {Hebbian} learning
  through spike-timing-dependent synaptic plasticity,'' \emph{Nature
  neuroscience}, vol.~3, no.~9, pp. 919--926, 2000.

\bibitem{guyonneau2005neurons}
R.~Guyonneau, R.~VanRullen, and S.~J. Thorpe, ``Neurons tune to the earliest
  spikes through {STDP},'' \emph{Neural Computation}, vol.~17, no.~4, pp.
  859--879, 2005.

\bibitem{masquelier2008spike}
T.~Masquelier, R.~Guyonneau, and S.~J. Thorpe, ``Spike timing dependent
  plasticity finds the start of repeating patterns in continuous spike
  trains,'' \emph{PloS one}, vol.~3, no.~1, p. e1377, 2008.

\bibitem{masquelier2018optimal}
T.~Masquelier and S.~R. Kheradpisheh, ``Optimal localist and distributed coding
  of spatiotemporal spike patterns through stdp and coincidence detection,''
  \emph{arXiv preprint arXiv:1803.00447}, p.~99, 2018.

\bibitem{masquelier2009competitive}
T.~Masquelier, R.~Guyonneau, and S.~J. Thorpe, ``Competitive {STDP}-based spike
  pattern learning,'' \emph{Neural computation}, vol.~21, no.~5, pp.
  1259--1276, 2009.

\bibitem{masquelier2010}
T.~Masquelier and S.~J. Thorpe, ``Learning to recognize objects using waves of
  spikes and spike timing-dependent plasticity,'' in \emph{Neural Networks
  (IJCNN), The 2010 International Joint Conference on}.\hskip 1em plus 0.5em
  minus 0.4em\relax IEEE, 2010, pp. 1--8.

\bibitem{tavanaei2016b}
A.~Tavanaei and A.~S. Maida, ``A spiking network that learns to extract spike
  signatures from speech signals,'' \emph{Neurocomputing}, vol. 240, pp.
  191--199, 2017.

\bibitem{kheradpisheh2016a}
S.~R. Kheradpisheh, M.~Ganjtabesh, and T.~Masquelier, ``Bio-inspired
  unsupervised learning of visual features leads to robust invariant object
  recognition,'' \emph{Neurocomputing}, vol. 205, pp. 382--392, 2016.

\bibitem{rao2002}
R.~P. Rao, B.~A. Olshausen, and M.~S. Lewicki, \emph{Probabilistic models of
  the brain: Perception and neural function}.\hskip 1em plus 0.5em minus
  0.4em\relax MIT press, 2002.

\bibitem{doya2007}
K.~Doya, \emph{Bayesian brain: Probabilistic approaches to neural
  coding}.\hskip 1em plus 0.5em minus 0.4em\relax MIT press, 2007.

\bibitem{mozer2008}
M.~C. Mozer, H.~Pashler, and H.~Homaei, ``Optimal predictions in everyday
  cognition: The wisdom of individuals or crowds?'' \emph{Cognitive Science},
  vol.~32, no.~7, pp. 1133--1147, 2008.

\bibitem{kording2004}
K.~P. K{\"o}rding and D.~M. Wolpert, ``Bayesian integration in sensorimotor
  learning,'' \emph{Nature}, vol. 427, no. 6971, pp. 244--247, 2004.

\bibitem{nessler2009}
B.~Nessler, M.~Pfeiffer, and W.~Maass, ``{STDP} enables spiking neurons to
  detect hidden causes of their inputs,'' in \emph{Advances in neural
  information processing systems}, 2009, pp. 1357--1365.

\bibitem{nessler2013}
B.~Nessler, M.~Pfeiffer, L.~Buesing, and W.~Maass, ``Bayesian computation
  emerges in generic cortical microcircuits through spike-timing-dependent
  plasticity,'' \emph{PLoS Comput Biol}, vol.~9, no.~4, p. e1003037, 2013.

\bibitem{klampfl2013}
S.~Klampfl and W.~Maass, ``Emergence of dynamic memory traces in cortical
  microcircuit models through {STDP},'' \emph{The Journal of Neuroscience},
  vol.~33, no.~28, pp. 11\,515--11\,529, 2013.

\bibitem{kappel2014}
D.~Kappel, B.~Nessler, and W.~Maass, ``{STDP} installs in winner-take-all
  circuits an online approximation to hidden {Markov} model learning,''
  \emph{PLoS Comput Biol}, vol.~10, no.~3, p. e1003511, 2014.

\bibitem{tavanaei2015a}
A.~Tavanaei and A.~S. Maida, ``Studying the interaction of a hidden {Markov}
  model with a {Bayesian} spiking neural network,'' in \emph{Machine Learning
  for Signal Processing (MLSP), 2015 IEEE 25th International Workshop
  on}.\hskip 1em plus 0.5em minus 0.4em\relax IEEE, 2015, pp. 1--6.

\bibitem{tavanaei2016a}
------, ``Training a hidden {Markov} model with a {Bayesian} spiking neural
  network,'' \emph{Journal of Signal Processing Systems}, pp. 1--10, 2016.

\bibitem{rezende2011}
D.~J. Rezende, D.~Wierstra, and W.~Gerstner, ``Variational learning for
  recurrent spiking networks,'' in \emph{Advances in Neural Information
  Processing Systems}, 2011, pp. 136--144.

\bibitem{kullback1951}
S.~Kullback and R.~A. Leibler, ``On information and sufficiency,'' \emph{The
  annals of mathematical statistics}, vol.~22, no.~1, pp. 79--86, 1951.

\bibitem{brea2011}
J.~Brea, W.~Senn, and J.-P. Pfister, ``Sequence learning with hidden units in
  spiking neural networks,'' in \emph{Advances in neural information processing
  systems}, 2011, pp. 1422--1430.

\bibitem{pecevski2016}
D.~Pecevski and W.~Maass, ``Learning probabilistic inference through {STDP},''
  \emph{eneuro}, pp. ENEURO--0048, 2016.

\bibitem{zemel2004}
R.~S. Zemel, R.~Natarajan, P.~Dayan, and Q.~J. Huys, ``Probabilistic
  computation in spiking populations,'' in \emph{Advances in neural information
  processing systems}, 2004, pp. 1609--1616.

\bibitem{buesing2011}
L.~Buesing, J.~Bill, B.~Nessler, and W.~Maass, ``Neural dynamics as sampling: a
  model for stochastic computation in recurrent networks of spiking neurons,''
  \emph{PLoS Comput Biol}, vol.~7, no.~11, p. e1002211, 2011.

\bibitem{Bishop1995a}
C.~M. Bishop, \emph{Neural Networks for Pattern Recognition}.\hskip 1em plus
  0.5em minus 0.4em\relax Oxford University Press, 1995.

\bibitem{Grossberg1987a}
S.~Grossberg, ``Competitive learning: From interactive activation to adaptive
  resonance,'' \emph{Cognitive Science}, vol.~11, no. 23-63, 1987.

\bibitem{Lillicrap2016a}
T.~P. Lillicrap, D.~Cownden, D.~B. Tweed, and C.~J. Akerman, ``Randome synaptic
  feedback weights support error backpropagation for deep learning,''
  \emph{Nature Communications}, pp. 1--10, 2016.

\bibitem{Zenke2018a}
F.~Zenke and S.~Ganguli, ``Superspike: Supervised learning in multi-layer
  spiking neural networks,'' \emph{Neural Computation}, vol.~30, no.~6, pp.
  1514--1541, 2017.

\bibitem{bohte2002}
S.~M. Bohte, J.~N. Kok, and H.~La~Poutre, ``Error-backpropagation in temporally
  encoded networks of spiking neurons,'' \emph{Neurocomputing}, vol.~48, no.~1,
  pp. 17--37, 2002.

\bibitem{booij2005}
O.~Booij and H.~tat Nguyen, ``A gradient descent rule for spiking neurons
  emitting multiple spikes,'' \emph{Information Processing Letters}, vol.~95,
  no.~6, pp. 552--558, 2005.

\bibitem{ghosh2009}
S.~Ghosh-Dastidar and H.~Adeli, ``A new supervised learning algorithm for
  multiple spiking neural networks with application in epilepsy and seizure
  detection,'' \emph{Neural Networks}, vol.~22, no.~10, pp. 1419--1431, 2009.

\bibitem{liu2017mt}
T.~Liu, Z.~Liu, F.~Lin, Y.~Jin, G.~Quan, and W.~Wen, ``Mt-spike: a multilayer
  time-based spiking neuromorphic architecture with temporal error
  backpropagation,'' in \emph{Proceedings of the 36th International Conference
  on Computer-Aided Design}.\hskip 1em plus 0.5em minus 0.4em\relax IEEE Press,
  2017, pp. 450--457.

\bibitem{mostafa2017supervised}
H.~Mostafa, ``Supervised learning based on temporal coding in spiking neural
  networks,'' \emph{IEEE transactions on neural networks and learning systems},
  pp. 1--9, 2017.

\bibitem{wu2017spatio}
Y.~Wu, L.~Deng, G.~Li, J.~Zhu, and L.~Shi, ``Spatio-temporal backpropagation
  for training high-performance spiking neural networks,'' \emph{arXiv preprint
  arXiv:1706.02609}, pp. 1--10, 2017.

\bibitem{ponulak2010}
F.~Ponulak and A.~Kasinski, ``Supervised learning in spiking neural networks
  with {ReSuMe}: sequence learning, classification, and spike shifting,''
  \emph{Neural Computation}, vol.~22, no.~2, pp. 467--510, 2010.

\bibitem{Kasinski2006a}
A.~Kasinski and F.~Ponulak, ``Comparison of supervised learning platforms for
  spike time coding in spiking neural networks,'' \emph{Int. J. Appl. Math.
  Comput. Sci}, vol.~16, no.~1, pp. 101--113, 2006.

\bibitem{florian2012}
R.~V. Florian, ``The chronotron: A neuron that learns to fire temporally
  precise spike patterns,'' \emph{PLOS One}, vol.~7, no.~8, p. e40233, 2012.

\bibitem{mohemmed2012}
A.~Mohemmed, S.~Schliebs, S.~Matsuda, and N.~Kasabov, ``Span: Spike pattern
  association neuron for learning spatio-temporal spike patterns,''
  \emph{International Journal of Neural Systems}, vol.~22, no.~04, p. 1250012,
  2012.

\bibitem{mohemmed2013}
------, ``Training spiking neural networks to associate spatio-temporal
  input-output spike patterns,'' \emph{Neurocomputing}, vol. 107, pp. 3--10,
  2013.

\bibitem{gutig2006}
R.~G{\"u}tig and H.~Sompolinsky, ``The tempotron: a neuron that learns spike
  timing-based decisions,'' \emph{Nature neuroscience}, vol.~9, no.~3, pp.
  420--428, 2006.

\bibitem{victor1997}
J.~D. Victor and K.~P. Purpura, ``Metric-space analysis of spike trains:
  theory, algorithms and application,'' \emph{Network: computation in neural
  systems}, vol.~8, no.~2, pp. 127--164, 1997.

\bibitem{tavanaei2017bp}
A.~Tavanaei and A.~S. Maida, ``{BP-STDP}: Approximating backpropagation using
  spike timing dependent plasticity,'' \emph{arXiv preprint arXiv:1711.04214},
  pp. 1--20, 2017.

\bibitem{pfister2006}
J.-P. Pfister, T.~Toyoizumi, D.~Barber, and W.~Gerstner, ``Optimal
  spike-timing-dependent plasticity for precise action potential firing in
  supervised learning,'' \emph{Neural computation}, vol.~18, no.~6, pp.
  1318--1348, 2006.

\bibitem{wang2014}
J.~Wang, A.~Belatreche, L.~Maguire, and T.~M. McGinnity, ``An online supervised
  learning method for spiking neural networks with adaptive structure,''
  \emph{Neurocomputing}, vol. 144, pp. 526--536, 2014.

\bibitem{tavanaei2015b}
A.~Tavanaei and A.~S. Maida, ``A minimal spiking neural network to rapidly
  train and classify handwritten digits in binary and 10-digit tasks,''
  \emph{International journal of advanced research in artificial intelligence},
  vol.~4, no.~7, pp. 1--8, 2015.

\bibitem{mozafari2017first}
M.~Mozafari, S.~R. Kheradpisheh, T.~Masquelier, A.~Nowzari-Dalini, and
  M.~Ganjtabesh, ``First-spike based visual categorization using
  reward-modulated stdp,'' \emph{IEEE Transactions on Neural Networks and
  Learning Systems, In Press}, pp. 1--24, 2018.

\bibitem{goodfellow2016deep}
I.~Goodfellow, Y.~Bengio, and A.~Courville, \emph{Deep learning}.\hskip 1em
  plus 0.5em minus 0.4em\relax MIT Press, 2016.

\bibitem{kheradpisheh2016deep}
S.~R. Kheradpisheh, M.~Ghodrati, M.~Ganjtabesh, and T.~Masquelier, ``Deep
  networks can resemble human feed-forward vision in invariant object
  recognition,'' \emph{Scientific reports}, vol.~6, p. 32672, 2016.

\bibitem{kheradpisheh2016humans}
------, ``Humans and deep networks largely agree on which kinds of variation
  make object recognition harder,'' \emph{Frontiers in Computational
  Neuroscience}, vol.~10, p.~92, 2016.

\bibitem{kasabov2014neucube}
N.~K. Kasabov, ``Neucube: A spiking neural network architecture for mapping,
  learning and understanding of spatio-temporal brain data,'' \emph{Neural
  Networks}, vol.~52, pp. 62--76, 2014.

\bibitem{wysoski2008}
S.~G. Wysoski, L.~Benuskova, and N.~Kasabov, ``Fast and adaptive network of
  spiking neurons for multi-view visual pattern recognition,''
  \emph{Neurocomputing}, vol.~71, no.~13, pp. 2563--2575, 2008.

\bibitem{brader2007learning}
J.~M. Brader, W.~Senn, and S.~Fusi, ``Learning real-world stimuli in a neural
  network with spike-driven synaptic dynamics,'' \emph{Neural computation},
  vol.~19, no.~11, pp. 2881--2912, 2007.

\bibitem{eliasmith2012large}
C.~Eliasmith, T.~C. Stewart, X.~Choo, T.~Bekolay, T.~DeWolf, Y.~Tang, and
  D.~Rasmussen, ``A large-scale model of the functioning brain,''
  \emph{Science}, vol. 338, no. 6111, pp. 1202--1205, 2012.

\bibitem{diehl2015unsupervised}
P.~U. Diehl and M.~Cook, ``Unsupervised learning of digit recognition using
  spike-timing-dependent plasticity,'' \emph{Frontiers in Computational
  Neuroscience}, vol.~9, pp. 1--9, 2015.

\bibitem{morrison2007spike}
A.~Morrison, A.~Aertsen, and M.~Diesmann, ``Spike-timing-dependent plasticity
  in balanced random networks,'' \emph{Neural Computation}, vol.~19, no.~6, pp.
  1437--1467, 2007.

\bibitem{pfister2006triplets}
J.-P. Pfister and W.~Gerstner, ``Triplets of spikes in a model of spike
  timing-dependent plasticity,'' \emph{Journal of Neuroscience}, vol.~26,
  no.~38, pp. 9673--9682, 2006.

\bibitem{lecun1998mnist}
Y.~LeCun, C.~Cortes, and C.~J. Burges, ``The {MNIST} database,'' \emph{URL
  http://yann. lecun. com/exdb/mnist}, 1998.

\bibitem{bengio2015towards}
Y.~Bengio, D.-H. Lee, J.~Bornschein, T.~Mesnard, and Z.~Lin, ``Towards
  biologically plausible deep learning,'' \emph{arXiv preprint
  arXiv:1502.04156}, pp. 1--10, 2015.

\bibitem{hinton2007backpropagation}
G.~Hinton, ``How to do backpropagation in a brain,'' in \emph{Invited talk at
  the NIPS'2007 Deep Learning Workshop}, vol. 656, 2007.

\bibitem{bengio2017stdp}
Y.~Bengio, T.~Mesnard, A.~Fischer, S.~Zhang, and Y.~Wu, ``{STDP}-compatible
  approximation of backpropagation in an energy-based model,'' \emph{Neural
  Computation}, pp. 555--577, 2017.

\bibitem{o2016deep}
P.~O'Connor and M.~Welling, ``Deep spiking networks,'' \emph{arXiv preprint
  arXiv:1602.08323}, pp. 1--16, 2016.

\bibitem{neftci2017event}
E.~O. Neftci, C.~Augustine, S.~Paul, and G.~Detorakis, ``Event-driven random
  back-propagation: Enabling neuromorphic deep learning machines,''
  \emph{Frontiers in Neuroscience}, vol.~11, p. 324, 2017.

\bibitem{querlioz2013immunity}
D.~Querlioz, O.~Bichler, P.~Dollfus, and C.~Gamrat, ``Immunity to device
  variations in a spiking neural network with memristive nanodevices,''
  \emph{IEEE Transactions on Nanotechnology}, vol.~12, no.~3, pp. 288--295,
  2013.

\bibitem{diehl2015}
P.~U. Diehl, D.~Neil, J.~Binas, M.~Cook, S.-C. Liu, and M.~Pfeiffer,
  ``Fast-classifying, high-accuracy spiking deep networks through weight and
  threshold balancing,'' in \emph{Neural Networks (IJCNN), 2015 International
  Joint Conference on}.\hskip 1em plus 0.5em minus 0.4em\relax IEEE, 2015, pp.
  1--8.

\bibitem{esser2015backpropagation}
S.~K. Esser, R.~Appuswamy, P.~Merolla, J.~V. Arthur, and D.~S. Modha,
  ``Backpropagation for energy-efficient neuromorphic computing,'' in
  \emph{Advances in Neural Information Processing Systems}, 2015, pp.
  1117--1125.

\bibitem{rueckauer2017conversion}
B.~Rueckauer, Y.~Hu, I.-A. Lungu, M.~Pfeiffer, and S.-C. Liu, ``Conversion of
  continuous-valued deep networks to efficient event-driven networks for image
  classification,'' \emph{Frontiers in Neuroscience}, vol.~11, p. 682, 2017.

\bibitem{stromatias2017event}
E.~Stromatias, M.~Soto, T.~Serrano-Gotarredona, and B.~Linares-Barranco, ``An
  event-driven classifier for spiking neural networks fed with synthetic or
  dynamic vision sensor data,'' \emph{Frontiers in Neuroscience}, vol.~11, p.
  350, 2017.

\bibitem{neil2016learning}
D.~Neil, M.~Pfeiffer, and S.-C. Liu, ``Learning to be efficient: Algorithms for
  training low-latency, low-compute deep spiking neural networks,'' in
  \emph{Proceedings of the 31st Annual ACM Symposium on Applied
  Computing}.\hskip 1em plus 0.5em minus 0.4em\relax ACM, 2016, pp. 293--298.

\bibitem{rawat2017deep}
W.~Rawat and Z.~Wang, ``Deep convolutional neural networks for image
  classification: A comprehensive review,'' \emph{Neural Computation}, pp.
  2352--2449, 2017.

\bibitem{oquab2014learning}
M.~Oquab, L.~Bottou, I.~Laptev, and J.~Sivic, ``Learning and transferring
  mid-level image representations using convolutional neural networks,'' in
  \emph{Proceedings of the IEEE conference on computer vision and pattern
  recognition}, 2014, pp. 1717--1724.

\bibitem{simonyan2014very}
K.~Simonyan and A.~Zisserman, ``Very deep convolutional networks for
  large-scale image recognition,'' \emph{arXiv preprint arXiv:1409.1556}, pp.
  1--14, 2014.

\bibitem{sainath2013deep}
T.~N. Sainath, A.-r. Mohamed, B.~Kingsbury, and B.~Ramabhadran, ``Deep
  convolutional neural networks for {LVCSR},'' in \emph{Acoustics, speech and
  signal processing (ICASSP), 2013 IEEE international conference on}.\hskip 1em
  plus 0.5em minus 0.4em\relax IEEE, 2013, pp. 8614--8618.

\bibitem{abdel2012applying}
O.~Abdel-Hamid, A.-r. Mohamed, H.~Jiang, and G.~Penn, ``Applying convolutional
  neural networks concepts to hybrid nn-hmm model for speech recognition,'' in
  \emph{Acoustics, Speech and Signal Processing (ICASSP), 2012 IEEE
  International Conference on}.\hskip 1em plus 0.5em minus 0.4em\relax IEEE,
  2012, pp. 4277--4280.

\bibitem{abdel2013exploring}
O.~Abdel-Hamid, L.~Deng, and D.~Yu, ``Exploring convolutional neural network
  structures and optimization techniques for speech recognition.'' in
  \emph{Interspeech}, 2013, pp. 3366--3370.

\bibitem{zeng2016convolutional}
H.~Zeng, M.~D. Edwards, G.~Liu, and D.~K. Gifford, ``Convolutional neural
  network architectures for predicting {DNA}--protein binding,''
  \emph{Bioinformatics}, vol.~32, no.~12, pp. i121--i127, 2016.

\bibitem{quang2016danq}
D.~Quang and X.~Xie, ``Danq: a hybrid convolutional and recurrent deep neural
  network for quantifying the function of {DNA} sequences,'' \emph{Nucleic
  acids research}, vol.~44, no.~11, pp. e107--e107, 2016.

\bibitem{tavanaei2016towards}
A.~Tavanaei, A.~S. Maida, A.~Kaniymattam, and R.~Loganantharaj, ``Towards
  recognition of protein function based on its structure using deep
  convolutional networks,'' in \emph{Bioinformatics and Biomedicine (BIBM),
  2016 IEEE International Conference on}.\hskip 1em plus 0.5em minus
  0.4em\relax IEEE, 2016, pp. 145--149.

\bibitem{ronneberger2015u}
O.~Ronneberger, P.~Fischer, and T.~Brox, ``U-net: Convolutional networks for
  biomedical image segmentation,'' in \emph{International Conference on Medical
  Image Computing and Computer-Assisted Intervention}.\hskip 1em plus 0.5em
  minus 0.4em\relax Springer, 2015, pp. 234--241.

\bibitem{lecun1998gradient}
Y.~LeCun, L.~Bottou, Y.~Bengio, and P.~Haffner, ``Gradient-based learning
  applied to document recognition,'' \emph{Proceedings of the IEEE}, vol.~86,
  no.~11, pp. 2278--2324, 1998.

\bibitem{lecun2015lenet}
Y.~LeCun \emph{et~al.}, ``Lenet-5, convolutional neural networks,'' \emph{URL:
  http://yann. lecun. com/exdb/lenet}, 2015.

\bibitem{szegedy2015going}
C.~Szegedy, W.~Liu, Y.~Jia, P.~Sermanet, S.~Reed, D.~Anguelov, D.~Erhan,
  V.~Vanhoucke, A.~Rabinovich \emph{et~al.}, ``Going deeper with
  convolutions.''\hskip 1em plus 0.5em minus 0.4em\relax CVPR, 2015.

\bibitem{marcelja1980mathematical}
S.~Mar{\^c}elja, ``Mathematical description of the responses of simple cortical
  cells,'' \emph{JOSA}, vol.~70, no.~11, pp. 1297--1300, 1980.

\bibitem{hubel1959receptive}
D.~H. Hubel and T.~N. Wiesel, ``Receptive fields of single neurones in the
  cat's striate cortex,'' \emph{The Journal of physiology}, vol. 148, no.~3,
  pp. 574--591, 1959.

\bibitem{hubel1962receptive}
------, ``Receptive fields, binocular interaction and functional architecture
  in the cat's visual cortex,'' \emph{The Journal of physiology}, vol. 160,
  no.~1, pp. 106--154, 1962.

\bibitem{foldiak1990forming}
P.~F{\"o}ldiak, ``Forming sparse representations by local anti-hebbian
  learning,'' \emph{Biological cybernetics}, vol.~64, no.~2, pp. 165--170,
  1990.

\bibitem{olshausen1996emergence}
B.~A. Olshausen \emph{et~al.}, ``Emergence of simple-cell receptive field
  properties by learning a sparse code for natural images,'' \emph{Nature},
  vol. 381, no. 6583, pp. 607--609, 1996.

\bibitem{bell1997independent}
A.~J. Bell and T.~J. Sejnowski, ``The ``independent components'' of natural
  scenes are edge filters,'' \emph{Vision research}, vol.~37, no.~23, pp.
  3327--3338, 1997.

\bibitem{rehn2007network}
M.~Rehn and F.~T. Sommer, ``A network that uses few active neurones to code
  visual input predicts the diverse shapes of cortical receptive fields,''
  \emph{Journal of computational neuroscience}, vol.~22, no.~2, pp. 135--146,
  2007.

\bibitem{zylberberg2011sparse}
J.~Zylberberg, J.~T. Murphy, and M.~R. DeWeese, ``A sparse coding model with
  synaptically local plasticity and spiking neurons can account for the diverse
  shapes of {V1} simple cell receptive fields,'' \emph{PLoS Comput Biol},
  vol.~7, no.~10, p. e1002250, 2011.

\bibitem{king2013inhibitory}
P.~D. King, J.~Zylberberg, and M.~R. DeWeese, ``Inhibitory interneurons
  decorrelate excitatory cells to drive sparse code formation in a spiking
  model of {V1},'' \emph{Journal of Neuroscience}, vol.~33, no.~13, pp.
  5475--5485, 2013.

\bibitem{savin2010independent}
C.~Savin, P.~Joshi, and J.~Triesch, ``Independent component analysis in spiking
  neurons,'' \emph{PLoS Comput Biol}, vol.~6, no.~4, p. e1000757, 2010.

\bibitem{burbank2015mirrored}
K.~S. Burbank, ``Mirrored {STDP} implements autoencoder learning in a network
  of spiking neurons,'' \emph{PLoS Comput Biol}, vol.~11, no.~12, p. e1004566,
  2015.

\bibitem{zhao2015feedforward}
B.~Zhao, R.~Ding, S.~Chen, B.~Linares-Barranco, and H.~Tang, ``Feedforward
  categorization on {AER} motion events using cortex-like features in a spiking
  neural network,'' \emph{IEEE transactions on neural networks and learning
  systems}, vol.~26, no.~9, pp. 1963--1978, 2015.

\bibitem{kheradpisheh2016stdp}
S.~R. Kheradpisheh, M.~Ganjtabesh, S.~J. Thorpe, and T.~Masquelier,
  ``Stdp-based spiking deep convolutional neural networks for object
  recognition,'' \emph{Neural Networks}, vol.~99, pp. 56--67, 2017.

\bibitem{tavanaei2016bio}
A.~Tavanaei and A.~S. Maida, ``Bio-inspired spiking convolutional neural
  network using layer-wise sparse coding and {STDP} learning,'' \emph{arXiv
  preprint arXiv:1611.03000}, pp. 1--16, 2016.

\bibitem{tavanaei2017multi}
------, ``Multi-layer unsupervised learning in a spiking convolutional neural
  network,'' in \emph{Neural Networks (IJCNN), 2017 International Joint
  Conference on}.\hskip 1em plus 0.5em minus 0.4em\relax IEEE, 2017, pp.
  81--88.

\bibitem{panda2016unsupervised}
P.~Panda and K.~Roy, ``Unsupervised regenerative learning of hierarchical
  features in spiking deep networks for object recognition,'' in
  \emph{International Conference on Neural Networks (IJCNN)}.\hskip 1em plus
  0.5em minus 0.4em\relax IEEE, 2016, pp. 299--306.

\bibitem{tavanaei2018training}
A.~Tavanaei, Z.~Kirby, and A.~S. Maida, ``Training spiking {ConvNets} by {STDP}
  and gradient descent,'' in \emph{Neural Networks (IJCNN), The 2018
  International Joint Conference on}.\hskip 1em plus 0.5em minus 0.4em\relax
  IEEE, 2018, pp. 1--8.

\bibitem{anwani2015normad}
N.~Anwani and B.~Rajendran, ``Normad-normalized approximate descent based
  supervised learning rule for spiking neurons,'' in \emph{2015 International
  Joint Conference on Neural Networks (IJCNN)}.\hskip 1em plus 0.5em minus
  0.4em\relax IEEE, 2015, pp. 1--8.

\bibitem{krizhevsky2009learning}
A.~Krizhevsky and G.~Hinton, ``Learning multiple layers of features from tiny
  images,'' pp. 1--60, 2009.

\bibitem{hunsberger2015spiking}
E.~Hunsberger and C.~Eliasmith, ``Spiking deep networks with {LIF} neurons,''
  \emph{arXiv preprint arXiv:1510.08829}, pp. 1--9, 2015.

\bibitem{hunsberger2016training}
------, ``Training spiking deep networks for neuromorphic hardware,''
  \emph{arXiv preprint arXiv:1611.05141}, 2016.

\bibitem{neil2016effective}
D.~Neil and S.-C. Liu, ``Effective sensor fusion with event-based sensors and
  deep network architectures,'' in \emph{Circuits and Systems (ISCAS), 2016
  IEEE International Symposium on}.\hskip 1em plus 0.5em minus 0.4em\relax
  IEEE, 2016, pp. 2282--2285.

\bibitem{indiveri2015neuromorphic}
G.~Indiveri, F.~Corradi, and N.~Qiao, ``Neuromorphic architectures for spiking
  deep neural networks,'' in \emph{Electron Devices Meeting (IEDM), 2015 IEEE
  International}.\hskip 1em plus 0.5em minus 0.4em\relax IEEE, 2015, pp. 4--2.

\bibitem{garbin2014variability}
D.~Garbin, O.~Bichler, E.~Vianello, Q.~Rafhay, C.~Gamrat, L.~Perniola,
  G.~Ghibaudo, and B.~DeSalvo, ``Variability-tolerant convolutional neural
  network for pattern recognition applications based on oxram synapses,'' in
  \emph{Electron Devices Meeting (IEDM), 2014 IEEE International}.\hskip 1em
  plus 0.5em minus 0.4em\relax IEEE, 2014, pp. 28--4.

\bibitem{esser2016convolutional}
S.~K. Esser, P.~A. Merolla, J.~V. Arthur, A.~S. Cassidy, R.~Appuswamy,
  A.~Andreopoulos, D.~J. Berg, J.~L. McKinstry, T.~Melano, D.~R. Barch
  \emph{et~al.}, ``Convolutional networks for fast, energy-efficient
  neuromorphic computing,'' \emph{Proceedings of the National Academy of
  Sciences}, p. 201604850, 2016.

\bibitem{cao2015}
Y.~Cao, Y.~Chen, and D.~Khosla, ``Spiking deep convolutional neural networks
  for energy-efficient object recognition,'' \emph{International Journal of
  Computer Vision}, vol. 113, no.~1, pp. 54--66, 2015.

\bibitem{rueckauer2016theory}
B.~Rueckauer, I.-A. Lungu, Y.~Hu, and M.~Pfeiffer, ``Theory and tools for the
  conversion of analog to spiking convolutional neural networks,'' \emph{arXiv
  preprint arXiv:1612.04052}, pp. 1--9, 2016.

\bibitem{deng2009imagenet}
J.~Deng, W.~Dong, R.~Socher, L.-J. Li, K.~Li, and L.~Fei-Fei, ``Imagenet: A
  large-scale hierarchical image database,'' in \emph{Computer Vision and
  Pattern Recognition, 2009. CVPR 2009. IEEE Conference on}.\hskip 1em plus
  0.5em minus 0.4em\relax IEEE, 2009, pp. 248--255.

\bibitem{hinton2006}
G.~E. Hinton, S.~Osindero, and Y.-W. Teh, ``A fast learning algorithm for deep
  belief nets,'' \emph{Neural computation}, vol.~18, no.~7, pp. 1527--1554,
  2006.

\bibitem{salama2010deep}
M.~A. Salama, A.~E. Hassanien, and A.~A. Fahmy, ``Deep belief network for
  clustering and classification of a continuous data,'' in \emph{Signal
  Processing and Information Technology (ISSPIT), 2010 IEEE International
  Symposium on}.\hskip 1em plus 0.5em minus 0.4em\relax IEEE, 2010, pp.
  473--477.

\bibitem{le2008representational}
N.~Le~Roux and Y.~Bengio, ``Representational power of restricted boltzmann
  machines and deep belief networks,'' \emph{Neural computation}, vol.~20,
  no.~6, pp. 1631--1649, 2008.

\bibitem{le2010deep}
------, ``Deep belief networks are compact universal approximators,''
  \emph{Neural computation}, vol.~22, no.~8, pp. 2192--2207, 2010.

\bibitem{lee2008sparse}
H.~Lee, C.~Ekanadham, and A.~Y. Ng, ``Sparse deep belief net model for visual
  area {V2},'' in \emph{Advances in neural information processing systems},
  2008, pp. 873--880.

\bibitem{lee2011unsupervised}
H.~Lee, R.~Grosse, R.~Ranganath, and A.~Y. Ng, ``Unsupervised learning of
  hierarchical representations with convolutional deep belief networks,''
  \emph{Communications of the ACM}, vol.~54, no.~10, pp. 95--103, 2011.

\bibitem{krizhevsky2010convolutional}
A.~Krizhevsky and G.~Hinton, ``Convolutional deep belief networks on
  {CIFAR}-10,'' \emph{Unpublished manuscript}, vol.~40, 2010.

\bibitem{susskind2008generating}
J.~M. Susskind, G.~E. Hinton, J.~R. Movellan, and A.~K. Anderson, ``Generating
  facial expressions with deep belief nets,'' in \emph{Affective
  Computing}.\hskip 1em plus 0.5em minus 0.4em\relax InTech, 2008.

\bibitem{mleczko2015rough}
W.~K. Mleczko, T.~Kapu{\'s}ci{\'n}ski, and R.~K. Nowicki, ``Rough deep belief
  network-application to incomplete handwritten digits pattern
  classification,'' in \emph{International Conference on Information and
  Software Technologies}.\hskip 1em plus 0.5em minus 0.4em\relax Springer,
  2015, pp. 400--411.

\bibitem{liu2014facial}
P.~Liu, S.~Han, Z.~Meng, and Y.~Tong, ``Facial expression recognition via a
  boosted deep belief network,'' in \emph{Proceedings of the IEEE Conference on
  Computer Vision and Pattern Recognition}, 2014, pp. 1805--1812.

\bibitem{lee2009unsupervised}
H.~Lee, P.~Pham, Y.~Largman, and A.~Y. Ng, ``Unsupervised feature learning for
  audio classification using convolutional deep belief networks,'' in
  \emph{Advances in neural information processing systems}, 2009, pp.
  1096--1104.

\bibitem{kang2013multi}
S.~Kang, X.~Qian, and H.~Meng, ``Multi-distribution deep belief network for
  speech synthesis,'' in \emph{Acoustics, Speech and Signal Processing
  (ICASSP), 2013 IEEE International Conference on}.\hskip 1em plus 0.5em minus
  0.4em\relax IEEE, 2013, pp. 8012--8016.

\bibitem{mohamed2009deep}
A.-r. Mohamed, G.~Dahl, and G.~Hinton, ``Deep belief networks for phone
  recognition,'' in \emph{Nips workshop on deep learning for speech recognition
  and related applications}, vol.~1, no.~9, 2009, p.~39.

\bibitem{hamel2010learning}
P.~Hamel and D.~Eck, ``Learning features from music audio with deep belief
  networks.'' in \emph{ISMIR}, vol.~10.\hskip 1em plus 0.5em minus 0.4em\relax
  Utrecht, The Netherlands, 2010, pp. 339--344.

\bibitem{mohamed2012acoustic}
A.-r. Mohamed, G.~E. Dahl, and G.~Hinton, ``Acoustic modeling using deep belief
  networks,'' \emph{IEEE Transactions on Audio, Speech, and Language
  Processing}, vol.~20, no.~1, pp. 14--22, 2012.

\bibitem{kuremoto2014time}
T.~Kuremoto, S.~Kimura, K.~Kobayashi, and M.~Obayashi, ``Time series
  forecasting using a deep belief network with restricted boltzmann machines,''
  \emph{Neurocomputing}, vol. 137, pp. 47--56, 2014.

\bibitem{jo2015improving}
T.~Jo, J.~Hou, J.~Eickholt, and J.~Cheng, ``Improving protein fold recognition
  by deep learning networks,'' \emph{Scientific Reports}, vol.~5, p. 17573,
  2015.

\bibitem{neftci2014}
E.~Neftci, S.~Das, B.~Pedroni, K.~Kreutz-Delgado, and G.~Cauwenberghs,
  ``Event-driven contrastive divergence for spiking neuromorphic systems,''
  \emph{Frontiers in Neuroscience}, vol.~8, pp. 1--14, 2014.

\bibitem{o2013real}
P.~O'Connor, D.~Neil, S.-C. Liu, T.~Delbruck, and M.~Pfeiffer, ``Real-time
  classification and sensor fusion with a spiking deep belief network,''
  \emph{Frontiers in Neuroscience}, vol.~7, pp. 1--13, 2013.

\bibitem{stromatias2015robustness}
E.~Stromatias, D.~Neil, M.~Pfeiffer, F.~Galluppi, S.~B. Furber, and S.-C. Liu,
  ``Robustness of spiking deep belief networks to noise and reduced bit
  precision of neuro-inspired hardware platforms,'' \emph{Frontiers in
  Neuroscience}, vol.~9, pp. 1--14, 2015.

\bibitem{stromatias2015scalable}
E.~Stromatias, D.~Neil, F.~Galluppi, M.~Pfeiffer, S.-C. Liu, and S.~Furber,
  ``Scalable energy-efficient, low-latency implementations of trained spiking
  deep belief networks on spinnaker,'' in \emph{Neural Networks (IJCNN), 2015
  International Joint Conference on}.\hskip 1em plus 0.5em minus 0.4em\relax
  IEEE, 2015, pp. 1--8.

\bibitem{merolla2011digital}
P.~Merolla, J.~Arthur, F.~Akopyan, N.~Imam, R.~Manohar, and D.~S. Modha, ``A
  digital neurosynaptic core using embedded crossbar memory with 45pj per spike
  in 45nm,'' in \emph{Custom Integrated Circuits Conference (CICC), 2011
  IEEE}.\hskip 1em plus 0.5em minus 0.4em\relax IEEE, 2011, pp. 1--4.

\bibitem{neil2014minitaur}
D.~Neil and S.-C. Liu, ``Minitaur, an event-driven {FPGA}-based spiking network
  accelerator,'' \emph{IEEE Transactions on Very Large Scale Integration (VLSI)
  Systems}, vol.~22, no.~12, pp. 2621--2628, 2014.

\bibitem{barra2012equivalence}
A.~Barra, A.~Bernacchia, E.~Santucci, and P.~Contucci, ``On the equivalence of
  hopfield networks and boltzmann machines,'' \emph{Neural Networks}, vol.~34,
  pp. 1--9, 2012.

\bibitem{hopfield1982neural}
J.~J. Hopfield, ``Neural networks and physical systems with emergent collective
  computational abilities,'' \emph{Proceedings of the national academy of
  sciences}, vol.~79, no.~8, pp. 2554--2558, 1982.

\bibitem{Hertz1991a}
J.~Hertz, A.~Krogh, and R.~G. Palmer, \emph{Introduction to the Theory of
  Neural Computation}.\hskip 1em plus 0.5em minus 0.4em\relax Addison-Wesley,
  1991.

\bibitem{barra2017phase}
A.~Barra, G.~Genovese, P.~Sollich, and D.~Tantari, ``Phase transitions in
  restricted boltzmann machines with generic priors,'' \emph{Physical Review
  E}, vol.~96, no.~4, p. 042156, 2017.

\bibitem{barra2018phase}
------, ``Phase diagram of restricted boltzmann machines and generalized
  hopfield networks with arbitrary priors,'' \emph{Physical Review E}, vol.~97,
  no.~2, p. 022310, 2018.

\bibitem{tubiana2017emergence}
J.~Tubiana and R.~Monasson, ``Emergence of compositional representations in
  restricted boltzmann machines,'' \emph{Physical Review Letters}, vol. 118,
  no.~13, p. 138301, 2017.

\bibitem{bengio1994learning}
Y.~Bengio, P.~Simard, and P.~Frasconi, ``Learning long-term dependencies with
  gradient descent is difficult,'' \emph{IEEE Transactions on Neural Networks},
  vol.~5, no.~2, pp. 157--166, 1994.

\bibitem{hochreiter1997long}
S.~Hochreiter and J.~Schmidhuber, ``Long short-term memory,'' \emph{Neural
  Computation}, vol.~9, no.~8, pp. 1735--1780, 1997.

\bibitem{chung2014empirical}
J.~Chung, C.~Gulcehre, K.~Cho, and Y.~Bengio, ``Empirical evaluation of gated
  recurrent neural networks on sequence modeling,'' \emph{arXiv preprint
  arXiv:1412.3555}, 2014.

\bibitem{shrestha2017spike}
A.~Shrestha, K.~Ahmed, Y.~Wang, D.~P. Widemann, A.~T. Moody, B.~C. Van~Essen,
  and Q.~Qiu, ``A spike-based long short-term memory on a neurosynaptic
  processor,'' in \emph{Computer-Aided Design (ICCAD), 2017 IEEE/ACM
  International Conference on}.\hskip 1em plus 0.5em minus 0.4em\relax IEEE,
  2017, pp. 631--637.

\bibitem{akopyan2015truenorth}
F.~Akopyan, J.~Sawada, A.~Cassidy, R.~Alvarez-Icaza, J.~Arthur, P.~Merolla,
  N.~Imam, Y.~Nakamura, P.~Datta, G.-J. Nam \emph{et~al.}, ``Truenorth: Design
  and tool flow of a 65 mw 1 million neuron programmable neurosynaptic chip,''
  \emph{IEEE Transactions on Computer-Aided Design of Integrated Circuits and
  Systems}, vol.~34, no.~10, pp. 1537--1557, 2015.

\bibitem{Neil2016cPhased}
D.~Neil, M.~Pfeiffer, and S.~Liu, ``Phased lstm: Accelerating neural network
  training for long or event-based sequences,'' in \emph{NIPS'16 Proceedings of
  the 30th International Conference on Neural Information Processing Systems},
  2016, pp. 3889--3897.

\bibitem{maass2002}
W.~Maass, T.~Natschl{\"a}ger, and H.~Markram, ``Real-time computing without
  stable states: A new framework for neural computation based on
  perturbations,'' \emph{Neural Computation}, vol.~14, no.~11, pp. 2531--2560,
  2002.

\bibitem{lukovsevivcius2009reservoir}
M.~Luko{\v{s}}evi{\v{c}}ius and H.~Jaeger, ``Reservoir computing approaches to
  recurrent neural network training,'' \emph{Computer Science Review}, vol.~3,
  no.~3, pp. 127--149, 2009.

\bibitem{lukovsevivcius2012reservoir}
M.~Luko{\v{s}}evi{\v{c}}ius, H.~Jaeger, and B.~Schrauwen, ``Reservoir computing
  trends,'' \emph{KI-K{\"u}nstliche Intelligenz}, vol.~26, no.~4, pp. 365--371,
  2012.

\bibitem{pyle2017spatiotemporal}
R.~Pyle and R.~Rosenbaum, ``Spatiotemporal dynamics and reliable computations
  in recurrent spiking neural networks,'' \emph{Physical Review Letters}, vol.
  118, no.~1, p. 018103, 2017.

\bibitem{paulun2018retinotopic}
L.~Paulun, A.~Wendt, and N.~K. Kasabov, ``A retinotopic spiking neural network
  system for accurate recognition of moving objects using {NeuCube} and dynamic
  vision sensors,'' \emph{Frontiers in Computational Neuroscience}, vol.~12,
  p.~42, 2018.

\bibitem{schliebs2011}
S.~Schliebs, H.~N.~A. Hamed, and N.~Kasabov, ``Reservoir-based evolving spiking
  neural network for spatio-temporal pattern recognition,'' in \emph{Neural
  Information Processing}.\hskip 1em plus 0.5em minus 0.4em\relax Springer,
  2011, pp. 160--168.

\bibitem{doborjeh2018modelling}
Z.~G. Doborjeh, N.~Kasabov, M.~G. Doborjeh, and A.~Sumich, ``Modelling
  peri-perceptual brain processes in a deep learning spiking neural network
  architecture,'' \emph{Scientific Reports}, vol.~8, no.~1, p. 8912, 2018.

\bibitem{bellec2018long}
G.~Bellec, D.~Salaj, A.~Subramoney, R.~Legenstein, and W.~Maass, ``Long
  short-term memory and learning-to-learn in networks of spiking neurons,''
  \emph{arXiv preprint arXiv:1803.09574}, 2018.

\bibitem{Lamb2016a}
A.~Lamb, A.~Goyal, Y.~Zhang, S.~Zhang, A.~Courville, and Y.~Bengio, ``Professor
  forcing: A new algorithm for training recurrent networks,'' \emph{arXiv
  preprint arXiv:1610.09038}, 2016.

\bibitem{Hochreiter2001a}
S.~Hochreiter, A.~S. Younger, and P.~R. Conwell, ``Learning to learn using
  gradient descent,'' in \emph{Intl Conf on Artificial Neural Networks}.\hskip
  1em plus 0.5em minus 0.4em\relax Springer, 2001, pp. 87--94.

\bibitem{costa2017cortical}
R.~Costa, I.~A. Assael, B.~Shillingford, N.~de~Freitas, and T.~Vogels,
  ``Cortical microcircuits as gated-recurrent neural networks,'' in
  \emph{Advances in Neural Information Processing Systems}, 2017, pp. 272--283.

\bibitem{orchard2015converting}
G.~Orchard, A.~Jayawant, G.~K. Cohen, and N.~Thakor, ``Converting static image
  datasets to spiking neuromorphic datasets using saccades,'' \emph{Frontiers
  in Neuroscience}, vol.~9, p. 437, 2015.

\end{thebibliography}

\end{document}